\documentclass[lettersize,journal]{IEEEtran}

\usepackage{cite}
\usepackage{amsmath,amssymb,amsfonts}
\usepackage{algorithmic}
\usepackage{graphicx}
\usepackage{textcomp}
\usepackage{xcolor}
\usepackage{hyperref}
\usepackage{array}
\usepackage{stfloats}
\usepackage{url}
\usepackage{verbatim}
\usepackage{xstring}
\usepackage{subfigure}
\usepackage{epsfig} 
\usepackage{times} 
\usepackage{mathtools}
\usepackage{mwe}
\usepackage{float}
\usepackage{adjustbox}
\usepackage{multirow}
\usepackage{rotating}

\begin{document}

\title{Design, Field Evaluation, and Traffic Analysis of a Competitive Autonomous Driving Model in a Congested Environment}

\author{Daegyu Lee$^{1*}$, 
        Hyunki Seong$^{1*}$, 
        Gyuree Kang$^1$, 
        Seungil Han$^2$, 
        D.Hyunchul Shim$^{1\dag}$, 
        and 
        Yoonjin Yoon$^{3\dag}$.
\thanks{$^{1}$School of Electrical Engineering, Korea Advanced Institute of Science and Technologies (KAIST), Daejeon, Republic of Korea
        {\texttt{\{lee.dk, hynkis, fingb20, hcshim\}@kaist.ac.kr}}}%
\thanks{$^{2}$Robotics Program, KAIST, Daejeon, Republic of Korea
        {\texttt{robotics@kaist.ac.kr}}}%
\thanks{$^{3}$Civil and Environmental Engineering, KAIST, Daejeon, Republic of Korea
        {\texttt{yoonjin@kaist.ac.kr}}}%
\thanks{$^{*}$Equally contributed}
\thanks{$^{\dag}$Co-corresponding author}
\thanks{This work is supported by the Institute of Information \& Communications Technology Planning Evaluation (IITP) grant funded by the Korean government (MSIT, 2021-0-00029).}
\thanks{This work has been submitted to the IEEE for possible publication. Copyright may be transferred without notice, after which this version may no longer be accessible.}
}

\markboth{Journal of \LaTeX\ Class Files,~Vol.~14, No.~8, August~2021}%
{Shell \MakeLowercase{\textit{et al.}}: A Sample Article Using IEEEtran.cls for IEEE Journals}


\maketitle

\begin{abstract}
Recently, numerous studies have investigated cooperative traffic systems using the communication among vehicle-to-everything (V2X).
Unfortunately, when multiple autonomous vehicles are deployed while exposed to communication failure, there might be a conflict of ideal conditions between various autonomous vehicles leading to adversarial situation on the roads.
In South Korea, virtual and real-world urban autonomous multi-vehicle races were held in March and November of 2021, respectively. 
During the competition, multiple vehicles were involved simultaneously, which required maneuvers such as overtaking low-speed vehicles, negotiating intersections, and obeying traffic laws.
In this study, we introduce a fully autonomous driving software stack to deploy a competitive driving model, which enabled us to win the urban autonomous multi-vehicle races. 
We evaluate module-based systems such as navigation, perception, and planning in real and virtual environments.
Additionally, an analysis of traffic is performed after collecting multiple vehicle position data over communication to gain additional insight into a multi-agent autonomous driving scenario.
Finally, we propose a method for analyzing traffic in order to compare the spatial distribution of multiple autonomous vehicles. 
We study the similarity distribution between each team's driving log data to determine the impact of competitive autonomous driving on the traffic environment.
\end{abstract}

\begin{IEEEkeywords}
Autonomous Vehicles, Traffic Information
\end{IEEEkeywords}

\section{Introduction}
\begin{figure}[t]
    \centering
    \includegraphics[width=0.9\columnwidth]{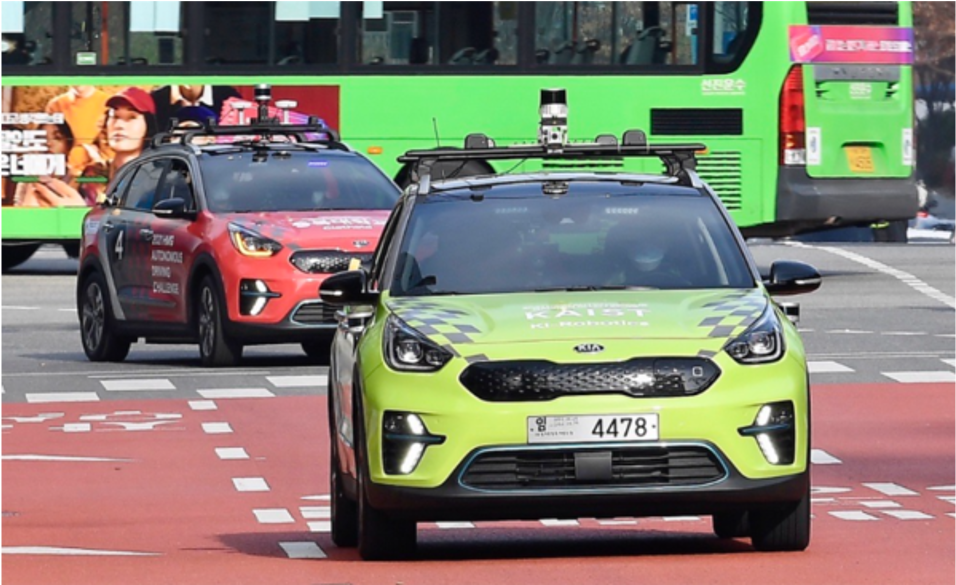}
    \caption[Niro final line]{The autonomous vehicle developed by the KAIST team competed in the Seoul competition held on November 29, 2021.}
    \label{fig:niro_final_line}
\end{figure}

\IEEEPARstart{R}{}ecent surge in vehicle-to-everything (V2X) based cooperative driving has demonstrated considerable improvements in effective traffic transition of autonomous vehicles, such as yielding and anticipatory slow-down \cite{morales2020automated}. 
When it comes to an urban environment, the requirements for effective autonomous driving becomes more complex. 
Due to the wide range of road conditions, including physical road geometry and communication reliability, more capabilities are necessary to maintain compliance with traffic law.
Especially, when vehicle drive in a congested urban environment, the capability to navigate in a competitive manner through the congested traffic plays the critical role.
However, when multiple autonomous vehicles are deployed, this competitive driving approach may lead to adversarial situations on the roads.
In this case, public exposure and acceptance is another critical factor \cite{fagnant2015preparing}. 
The negative impact of exposed communication failures or aggressive maneuvers can take months, if not years, to reverse.
\\
In this study, we propose a full stack of autonomous driving architecture that can effectively deal with various uncertainties in the complex urban environments. The architecture includes modular-level development of perception, localization, planning, and control. 
It requires only the traffic signal information through V2X, and provide efficient localization in the GPS-denied environment. 
A real-world experiment was carried out in the Hyundai Motor Group's biennial urban autonomous racing competition \cite{shim2015autonomous, jo2013overall, lee2016eurecar, jung2020v2x}. 
The competition is a multi-agent autonomous vehicle race on the congested streets in Seoul with the goal to minimize the transversal time while fully complying to the traffic law. 
The vehicle (Fig. \ref{fig:niro_final_line}) equipped with our architecture won the competition only with a simple feed on traffic signal information through vehicle-to-infrastructures(V2I) and no GPS. 
\\
The trajectory analysis revealed that our vehicle trajectory shares much lesser degrees of similarities among the five competitors in curve, intersection and straight maneuvers. 
Such observations not only validates the effectiveness of the proposed architecture, but also indicate that it can provide increased level of resilience in autonomous driving in urban environment. 
Furthermore, we can analyze how competitive driving approaches can result in adversarial situations on roads in multiple autonomous vehicle deployment scenarios.
\\
The remainder of this paper is organized as follows: Section \ref{sec:related} introduces previous related work. 
Section \ref{sec:system} presents an overview of our developed autonomous vehicle system. 
Section \ref{sec:methods} describes the development of robust localization, perception, motion planning, and behavior planning. 
Section \ref{sec:traffic} describes the traffic-level analysis of competitive driving models. 
In Section \ref{sec:experiments}, the experimental results of both a simulation and real-world application are presented. 
Finally, Section \ref{sec:conclusion} concludes this paper.

\section{Related Works}
\label{sec:related}
\subsection{Full-stack autonomy for self-driving}
In recent years, several studies on autonomous driving have been conducted.
Deep neural networks (DNNs) and reinforcement learning (RL)-based approaches have been used to deploy partial or full end-to-end autonomous driving \cite{bojarski2016end, codevilla2018end, xiao2020multimodal}. 
Furthermore, public-use studies \cite{kato2018autoware, fan2018baidu} have become open-source software projects for autonomous driving. 
Owing to these works, essential functions for autonomous driving are accessible to the public, and many individuals and organizations are able to utilize these contributed open sources \cite{raju2019performance}. 
Generally, a full-stack of autonomous technology is composed of primary modular layers: sensors, computing devices and their software interfaces, perceptions, plannings, and control \cite{pendleton2017perception, dai2020perception}. 
Owing to varying user requirements for autonomous stacks, integration of full-stack autonomy with public projects is essential for self-driving studies \cite{tokunaga2019idf, zang2022winning}. 
In this study, we propose a full-stack autonomy, covering localization, perception, and planning modules. 
Our full-stack autonomy utilizes state-of-the-art perception modules and the common concept of the control module.

\subsection{Autonomous racing competitions}
Competitions provide excellent motivation to quickly respond to and accelerate demands in robotics technology. 
In an effort to create the first fully autonomous ground vehicles, the Defense Advanced Research Projects Agency (DARPA) Grand Challenge was held to encourage the development of various robotics technologies \cite{thrun2006stanley,agha2021nebula, song2015darpa, lim2017robot}. 
F1Tenth is a competition to develop autonomous driving algorithms in a high-speed environment based on a race car that is one tenth the size of an actual car \cite{o2020f1tenth, patton2021neuromorphic}. 
This scale-model has an advantage in that it provides a handy platform for various experimental attempts related to autonomous driving research. 
Recently, an autonomous driving algorithm development contest in a high-speed environment using full-scale racing cars was held at Indianapolis Motor Speedway, and the same competition was held in Las Vegas shortly thereafter \cite{herrmann2020real, jung2021game, herrmann2020minimum, lee2022resilient}. 
Additionally, as part of our research project, our team also participated in a high-speed race. 

\begin{figure}[t!]
\centering
\includegraphics[width=1.0\columnwidth]{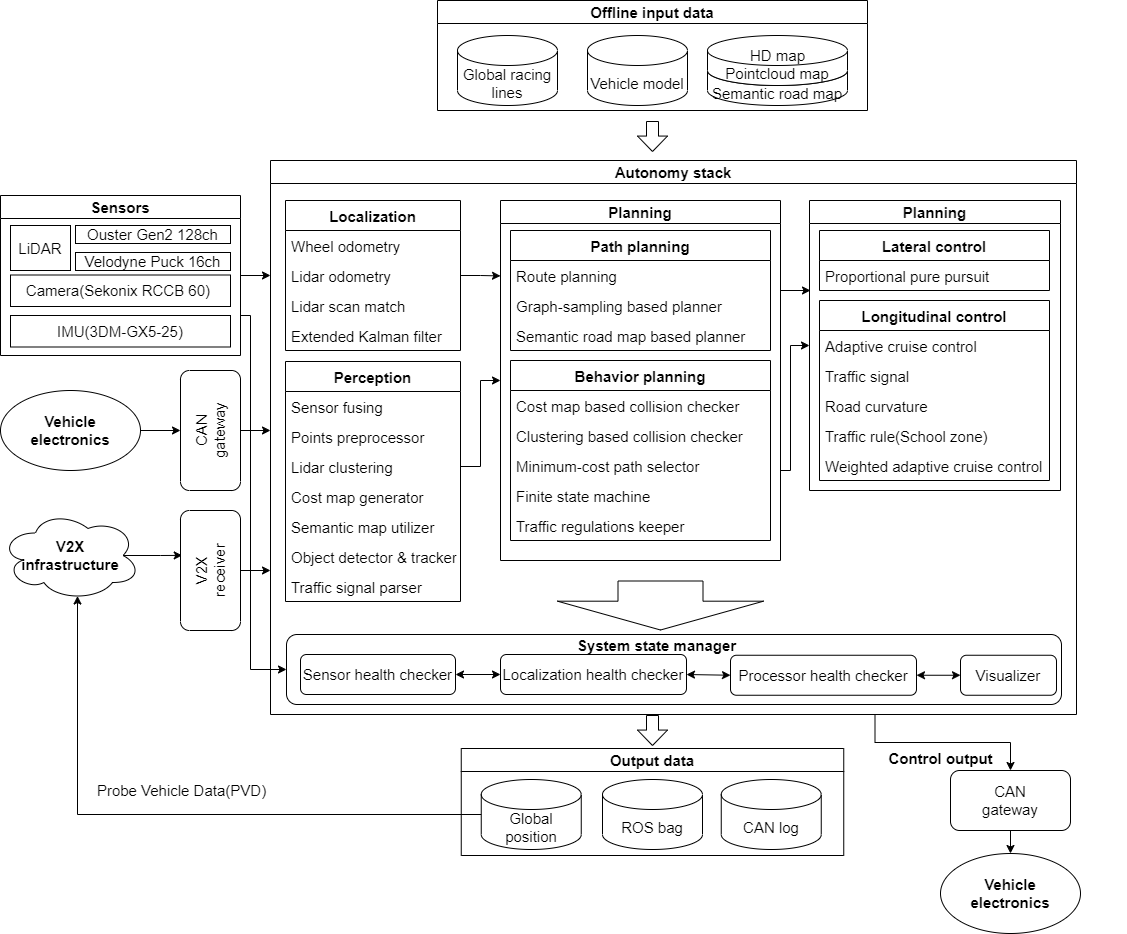}
\caption{The software system architecture for an autonomous vehicle.}
\label{fig:system_overview}
\end{figure}

\section{System Design}
\label{sec:system}
\subsection{Software}
Figure \ref{fig:system_overview} shows an overview of our autonomous driving system, composed of perception, planning, and control modules. 
An advantage of a modular autonomous pipeline is that we can contribute to the development of individual modulars with localization, planning, and perception algorithms. 
In addition, we manage the behavior of the autonomous vehicle based on the modular system, rather than an end-to-end autonomous pipeline. 
As a result, our vehicle estimates its position in the world with sensory data and executes planning and control algorithms sequentially. 
For perception, we developed a conservative multi-modal perception system utilizing an RGB camera, LiDAR point cloud, and semantic road map information. 
Moreover, the planning algorithm is designed hierarchically and consists of route planning, motion planning, and behavior planning. 
First, the route-planning algorithm determines the global path to arrive at the target position. 
Subsequently, the vehicle works to avoid obstacles encountered during the drive using a motion-primitives-based motion-planning algorithm. 
In low-speed traffic scenarios, competitive driving behavior planning allows the vehicle to overtake or change lanes to escape traffic congestion.
At the end of the autonomous stack, we control the electric vehicle directly over the controller area network (CAN) gateway.

\subsection{Hardware}
An Ouster OS2-128 LiDAR, two Velodyne Puck 16-channel LiDAR, two Sekonix SF3325-100 RGB cameras, and MicroStrain 3DM-GX5-IMU were used for the autonomous driving, with an industrial-grade computing platform (CPU: Intel Xeon, 8th-Gen Core processor, two of GPU: RTX 2020 Ti, 11 GB ), an Intel NUC (Intel i7-10710U Processor, 12 M cache, up to 4.70 GHz) and an Nvidia Drive PX 2 (AutoChauffeur) [CPU: 4x Denver, 8x Cortex A57, GPU: 2x Parker GPGPU (2x 2 SM Pascal, 512 CUDA cores) + 2x dedicated MXM modules] as the computing devices.
Using these three computing devices, our full-autonomous stack can obtain solutions for the localization, planning, and perception algorithms. 
Most of the algorithms were implemented in C++ and executed using the robot operating system (ROS) \cite{quigley2009ros} in Ubuntu 18.04 Linux. 
In addition, we employed a deep-learning-based algorithm with the Python script, and integrated inference results on the ROS system.

\section{Methods}
\label{sec:methods}
\subsection{Resilient localization}
The localization system is one of the most essential modules in the autonomous stack for real-world field testing. 
Unless a high-quality differential global positioning system (DGPS) sensor is used, global navigation satellite system (GNSS)-based localization quality is greatly deteriorated when passing near a building, especially in metropolitan areas due to a GPS multi-path problem or weak signal strength \cite{Kos}. 
However, some researchers have used a Kalman filter with a GPS \cite{Welch, Honghui,Reina,Hu, larsen1998location,larsen1999design,martinelli2003estimating}, and several studies have been conducted to enhance the localization system based on sensor fusion, while reducing the weight of the GPS \cite{ali2021real, 9328323}. 
Therefore, LiDAR or vision-sensor-based localization methods would be more ideal to estimate vehicle position. \\
In this study, we propose a novel localization system, without a GPS, utilizing only a pre-built 3-D point-cloud map, and RGB camera-based online lane detection. 
As a result of our previous work deploying robots in GPS-denied areas \cite{lee2021assistive}, we have built a localization system that enables location without a GNSS.
In a specific district-scale urban area, pre-built 3-D map-based localization \cite{Caselitz,Gu,Javanmardi, biber2003normal, ulacs20133d} can be utilized after constructing a 3-D point-cloud map from several milestone works \cite{LOAM, lego, lio, xu2021fast,nguyen2021liro}. 
One of the advantages of utilizing a pre-built map is that we can discover any map errors before deploying autonomous vehicles in the real world.

\subsubsection{Registration}
We define $W \subset \mathbb{R}^2$ as the map coordinates of the autonomous vehicle.
Let ${\hat{\mathbf{x}}}^{W}_{t} = [x_t, y_t, \theta_{t}] \in W$ be the estimated vehicle position $(x, y)$ with heading $\theta$ defined in the pre-built 3-D point-cloud map at time $t$. 
In addition, we define $B \subset \mathbb{R}^2$ as the vehicle body coordinates.
Here, the annotated values $B$ indicate information obtained from the origin of the vehicle body’s coordinates---i.e., the center point of the rear axle. 
We also define the voxel-filtered LiDAR points $\mathbf{z}^B_{t}= \{z_1^B, \dots, z_k^B\}$ at time $t$, where $z_i^B$ is one of the voxelized points from incoming LiDAR points, and the voxelized point-cloud map $\mathbf{M}^W = \{m_1^W, \dots, m_k^W\} $, where $m_i^W$ is one of the voxelized points of the original full-size point-cloud map.
As a result, the vehicle is considered to be located at the position ${\hat{\mathbf{x}}}^{W}_{t}$ in the 3-D map $\mathbf{M}^W$.
\\
We define our registration-based pose estimation problem  as minimizing the error $d_t$ between the voxelized LiDAR points $\mathbf{z}^W_{t}$ and $\mathbf{M}^W$  as follows: 
\begin{equation}
\begin{aligned}
    {d_t} = \arg\min_{d}\sum_{i =0}^{n}{\parallel(m_{i}^W - {z}_i^W)\parallel}_2,
\end{aligned}
\label{eq_fit}
\end{equation}
where $\mathbf{z}^W = \{z_{1}^W,\dots,z_{k}^W\}$ is transformed using the solution of (\ref{eq_fit}) which consists of transformation matrices $\mathbf{T}_t$ and $\mathbf{z}^B$ at time $t$. 
Therefore, the positioning problem can be considered as estimating $\mathbf{T}_t$ at time $t$ because the $\mathbf{T}_t \in SE(2)$ aligns $B$ to $W$. \\
Some studies \cite{sommer2020openmp, dellenbach2022ct} have implemented various algorithms to attain a much faster frequency than that using the conventional iterative closest point (ICP) variant algorithm \cite{chetverikov2002trimmed} to run a registration algorithm; by contrast, in this study, we focus on utilizing a high-density LiDAR point-cloud to register the 3-D points algorithm, except for the feature-based algorithm \cite{rusu2009fast,salti2014shot}. 
However, in the case of high-channel stacked up models, such as 64 or 128 channels (in this study, we installed an OS2-128 model), the LiDAR sensor driver publishes approximately 65 k and 130 k points every 0.1 $s$, for the 64- and 128-channel models, respectively.
Furthermore, the point size of our full-size point-cloud map, which is the general size of the district-scale area, is higher than 10,000 k points. 
Due to the characteristic of the ICP variant algorithm, depending on the number of inputs and target points, the computational burden severely increases, causing performance degradation.
However, the ICP variant algorithm takes advantage of the voxelization approach rather than the normal distributions transform (NDT)-based approach \cite{biber2003normal, koide2019portable}, leading to leverage on real-time computing performance \cite{koide2021voxelized}.
As a result, to perform an efficient registration algorithm up to the typical urban maximum speed of 50 kph, we employ the generalized iterative closest point (GICP) variant algorithm \cite{gicp, koide2021voxelized}, a registration algorithm for 3-D point-clouds, to model the point-represented environment as a Gaussian distribution, $\mathbf{z}^W_{t} \sim \mathcal{N} (\hat{z_i}, C^{z}_i)$,  $\mathbf{M}^W_{t} \sim \mathcal{N} (\hat{m_i}, C^{m}_i)$. Subsequently, the transformation error $d_i$ can be defined as
\begin{equation}
\begin{aligned}
    {d_i} = \hat{m_i} - \mathbf{T}_t\hat{z_i}.
\end{aligned}
\label{eq_trans_err}
\end{equation}
Thus, the $d_i$ distribution can be expressed as
\begin{equation}
\begin{aligned}
    {d_i} &\sim \mathcal{N} (\hat{m_i} - \mathbf{T}_t\hat{z_i}, C^{m}_i - \mathbf{T}_t^TC^{z}_i\mathbf{T}_t) \\
    &= \mathcal{N} (0, C^{m}_i - \mathbf{T}_t^TC^{z}_i\mathbf{T}_t).
\end{aligned}
\label{eq_trans_err_dist}
\end{equation}
Therefore, the vehicle position can be determined by calculating the $\mathbf{T}_t$ that maximizes the log likelihood of \eqref{eq_trans_err_dist}, such that
\begin{equation}
\begin{aligned}
    \mathbf{T}_t &= \arg\max_{\mathbf{T}_t}(\log(p(d_i))) \\
    &=\arg\min_{\mathbf{T}_t}(\sum_{i}d_i^T(C^{m}_i - \mathbf{T}_t^TC^{z}_i\mathbf{T}_t)d_i).
\end{aligned}
\label{eq_trans_likelihood}
\end{equation}
Equation \eqref{eq_trans_likelihood} can be re-defined such that the cost function $J(x)$ between the incoming LiDAR data $\mathbf{z}^B_{t}$ and $\mathbf{M}^W$ can be calculated as
\begin{equation}
\begin{aligned}
    {J(x)} = \arg\max_{J} \sum_{i=0}^{n}(-\frac{(\mathbf{M}^W-\mathbf{T}_{t}z_{i}^B)^{'}\Sigma^{-1}_{i}(\mathbf{M}^W-\mathbf{T}_{t}z_{i}^B)}{2}).
\end{aligned}
\label{eq_gicp}
\end{equation}
To achieve real-time operation, we utilize a Voxelized-GICP algorithm \cite{koide2021voxelized}, which extends the conventional GICP algorithm using the voxel-based association approach. \\
However, despite utilizing an enhanced registration algorithm, there is a still limitation in deploying an autonomous vehicle on the district-scale environment because the target point-cloud $\mathbf{M}^W$ size affects the calculation time for the solution of \eqref{eq_gicp}.
Therefore, we put a sliding-window of the point-cloud map as a limit to the size of $\mathbf{M}^W$ to accommodate the registration computing capacity. 
As the full-size $\mathbf{M}^W$ is not used in \eqref{eq_gicp}, we register only the points in a radius $\psi(v_{t})$ from the robot position $\hat{\mathbf{x}}_t$.
The surrounding radius increases proportionally to the velocity $v_{t}$ at time $t$,  
\begin{equation}
\begin{aligned}
    \delta^{\psi}_{i} = {\lVert m_i^W - \hat{\mathbf{x}}_t \rVert}_2,
\end{aligned}
\label{eq_closest_point}
\end{equation}
where $\delta^{\psi}_{i}$ indicates the distance from $\hat{\mathbf{x}}_t$ to the voxelized 3-D map points ${m}_i^W$. 
Therefore, the points in the sliding-window $\mathbf{M}_i^{\psi} = \{{m}_1^{\psi}, \dots, {m}_j^{\psi}\}$ replace the entire 3-D map points $\mathbf{M}^W$ in \eqref{eq_gicp}. Thus, 
\begin{equation}
    m_{i}^{\psi} = 
    \begin{cases*}
        {m}_{i}^W, 
        & if $\delta^{\psi}_i < \psi(v)$,\\
        none,                    & otherwise,
    \end{cases*}
\end{equation}
where point ${m}_i^{\psi}$ is selected from $\mathbf{M}^W$, which is determined from ${\delta}_i^{\psi} < \psi(v)$. 
Hence, we limit the registration target according to the LiDAR sensor frequency, which is approximately 10 Hz, without delay.
\begin{figure}[t]
\centering
    \includegraphics[width=1.0\columnwidth]{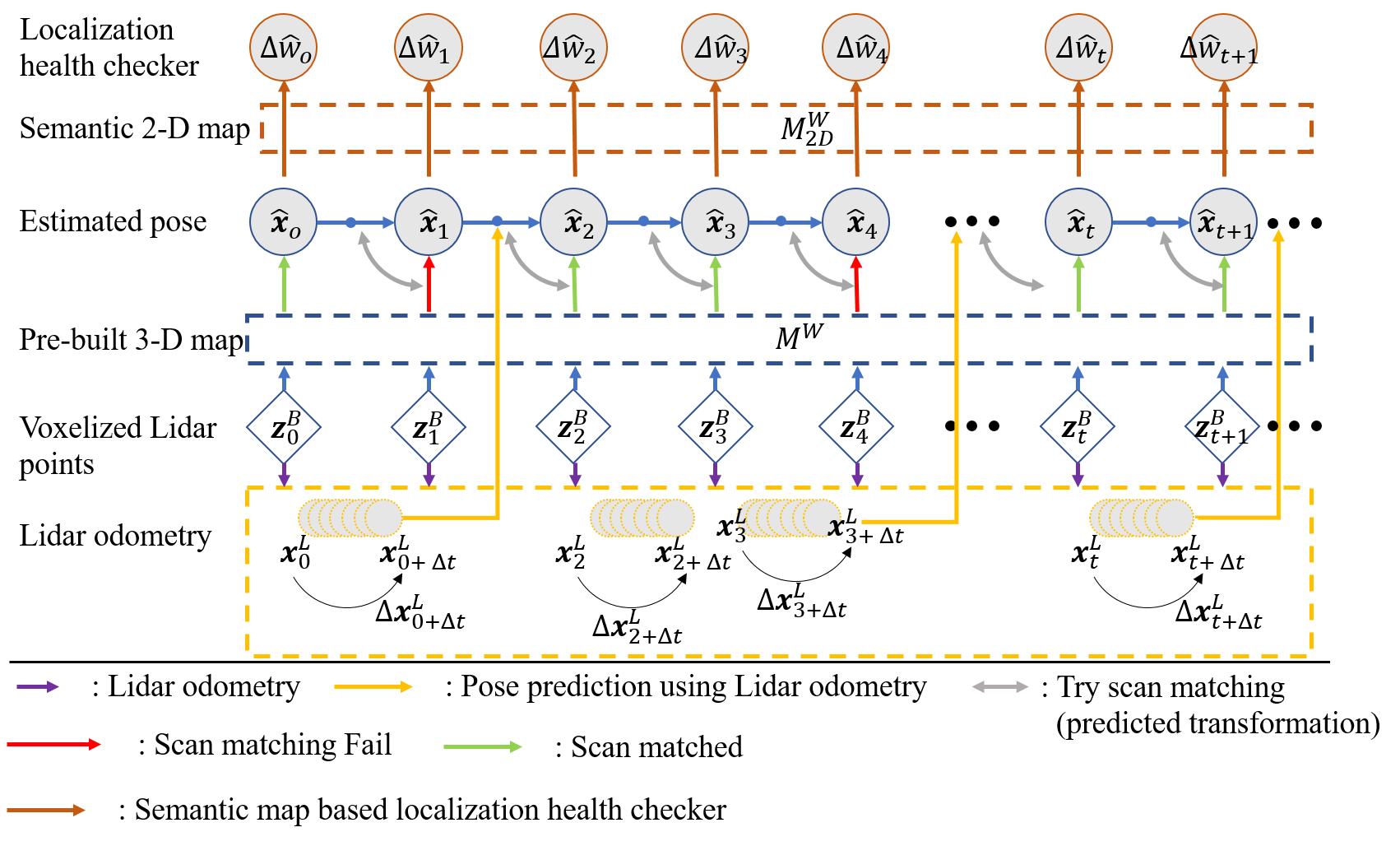}
    \caption[Robust localize]{Implementation of a robust localization system. LiDAR points are voxelized and used for registration and LiDAR odometry. As outlined by the red arrow, if the registration algorithm fails, prediction based on LiDAR odometry compensates the failure and enables the robot to implement the next step. Even though a vehicle moves dynamically, LiDAR-based odometry can calculate the pose accurately, thereby reducing the likelihood of scan-matching failures.}
    \label{fig:robust_localize}
\end{figure}

\subsubsection{Robustness}
The performance of this registration-based scan-matching algorithm is not always guaranteed, particularly for large-scale 3-D maps and dynamic driving scenarios.
Because an ICP-variant registration algorithm calculates the solution iteratively to find $\mathbf{T}_{t}$, the previous transformation matrix can be defined as $\mathbf{T}_{t-1}$.
Thus, we can define a scan-matching-based transition distance between $t$ and $t-1$ as ${\parallel\hat{\mathbf{x}}_{t} - \hat{\mathbf{x}}_{t-1}\parallel}_2$.
Because the ICP-variant registration algorithm starts to compute $\mathbf{T}_{t}$ from the initial input transform matrix ---i.e., in the registration algorithm, it is the initial guess transform matrix --- a rational guessing matrix input is crucial for an effective scan-matching performance.
In addition, there are parameters for the ICP-variant registration algorithm to constrain the maximum number of iterations $n_{max}$ and the searching corresponding-point-distance $\epsilon_{corres}$.
Because both $n_{max}$ and $\epsilon_{corres}$ affect the computing time, they are significant in maintaining the localization performance during the relatively high-speed driving.
As a result, the registration algorithm can be expected to solve the scan-matching points as
\begin{equation}
    \begin{aligned}
    n_{max} \times \epsilon_{corres} < {\parallel\hat{\mathbf{x}}_{t} - \hat{\mathbf{x}}_{t-1}\parallel}_2 .
    \end{aligned}
    \label{eq_fail_case}
\end{equation} 
However, in this study, we use LiDAR odometry for the robustness of the navigation system, without expecting scan-matching performance, while continuously increasing the value of $n_{max}$.
Therefore, we compute the LiDAR odometry $\mathbf{x}_{1:t}^{L} = [\mathbf{x}_1^{L}, \dots, \mathbf{x}_{t}^{L}] \in \mathbb{R}^2$ based on the LiDAR feature-based algorithm to obtain $\mathbf{x}_i^{L} = \{x_i^L, y_i^L, \theta_i^L\}$ \cite{LOAM, lego, lio, xu2021fast}.
We focus on LiDAR feature-based predictive transformation ${\Delta}\mathbf{x}_{t+{\Delta}t}^{L}$ to estimate the high-probabilistic guessing matrix for the ICP-variant registration algorithm as
\begin{equation}
\begin{aligned}
    {{\Delta}\mathbf{x}_{t+{\Delta}t}^{L}} = {{T^{L}}^{-1}(\mathbf{x}_{t+{\Delta}t}^{L} - \mathbf{x}_{t}^{L})},
\end{aligned}
\end{equation}
where $T^L(\mathbf{x})$ represents the rigid transformations from $\mathbf{x}_{1:t}^{L}$, and $T^{-1}$ represents an inverse transformation from $W$ to $B$.
As a result, we designed our robust localization system to detect when the registration algorithm fails due to the limitations of $n_{max}$ and $\epsilon_{corres}$.
\\
In the failure-case, we replace $\mathbf{T}_{t-1}$ with $\mathbf{T}_{t+{\Delta}t}$ ---i.e., in the normal case $\mathbf{T}_{t-1}$ is used for next-step registration where the predictive transformation $\mathbf{T}_{t+{\Delta}t}$ is computed from ${\hat{\mathbf{x}}_{t+{\Delta}t}}$, as follows:
\begin{equation}
\begin{aligned}
    {\hat{\mathbf{x}}_{t+{\Delta}t}} = {\hat{\mathbf{x}}_{t} + {\Delta}\mathbf{x}_{t+{\Delta}t}^{L}}.
\end{aligned}
\end{equation}
Finally, the robust localization algorithm is illustrated in Fig. \ref{fig:robust_localize}.

\subsubsection{Resilience}
\label{sec:resilience}
In this study, the use of high-quality DGPS is excluded for autonomous driving near high-rise buildings; instead, a localization algorithm is utilized based on a scan-matching algorithm using LiDAR sensors. 
Because stability of localization is one of the most crucial factors for operating an unmanned vehicle, the resilience of navigation algorithms is considered using RGB cameras, as well as LiDAR sensors. 
Our designed resilient-navigation system identifies a failure of LiDAR-based pose estimation and performs lane-detection-based control.
First, we compare our estimated pose $\hat{\mathbf{x}}_t$ with $M_{2D}^W$ to monitor whether our vehicle keeps to the center of the lane, which can be utilized as a localization health checker.
Second, switching to fail-safe mode, it conducts pose re-initialization itself based on road-marker recognition using an RGB camera. When the solution of \eqref{eq_fit} is higher than our scan-matching thresholds, the localization health checker gives an alert alarm, and our vehicle changes the control mode from navigation to lane detection.
If the solutions of the scan-matching algorithm and the lane-detection-based control are not feasible, this case is considered a localization failure and the vehicle stops. 
The resilient localization algorithm is depicted in Fig. \ref{fig:robust_localize}.

\subsection{Perception}

\subsubsection{V2X-enabled traffic signal}
A human-mimicking method, such as vision-perception-based traffic signal detection, is more intuitive than communication-based traffic signal recognition. 
However, V2X can be an excellent solution for stability when deploying vehicles in an urban area. 
In this study, we utilized V2I, which has a communication system between the vehicle and infrastructure \cite{jung2020v2x}. 
Specifically, we utilized TCP-based V2I in the infra-supported area by sending our pose $\hat{\mathbf{x}}_t$ and velocity to the infrastructure. 
According to standardized protocol \cite{kenney2011dedicated, park2011integrated}, we received the intersection name, signal state, and remaining signal time via Signal Phase and Timing (SPaT) messages.

\subsubsection{Object detection \& tracking algorithm}

Perception stacks is a critical module covering multi-discipline areas.
Recently, thanks to rapidly developing computer vision, several pioneering studies have enabled a vehicle to perceive the surrounding objects and environment. 
For safe driving in an urban environment, an autonomous vehicle must consider other vehicles, pedestrians, and static obstacles. 
To end this, in this study, we implemented a multi-modal sensor fusion algorithm utilizing camera and LiDAR units with public-use detection and tracking algorithms.
Therefore, objects $\{\mathcal{O}_i\}_{i \in [1:m]}$ are defined by detection modules. \\
For the optical camera, we employed the general bounding-box detection algorithm using front and rear cameras. 
To estimate the distance to the object, we implemented the azimuth-aware fusion algorithm, which derives the azimuth of the image bounding box and searches a corresponding LiDAR cluster $\mathcal{O}_{i}^{fusion} $. \\
For the LiDAR-only detection, we used the ResNet-based keypoint feature pyramid network \cite{li2020rtm3d} to convert point-cloud data into bird’s-eye-view images, and utilized the image-net-based approach.
Therefore, we can define a detected object $\mathcal{O}_{i}^{lidar}$ using the LiDAR-only method. 
Subsequently, the detected results $\{\mathcal{O}_{i}^{fusion}, \mathcal{O}_{j}^{lidar}\} \subset \{\mathcal{O}_i\}_{i \in [1:m]}$ obtained from the multi-modal and LiDAR-only methods are pipelined to the multi-object tracking(MOT) algorithm ---i.e., we employed the FastMOT \cite{yukai_yang_2020_4294717}, which guarantees real-time performance. 
Finally, the MOT algorithm assigns each object an identifying number.

\subsection{Route planning}
\label{sec:route_planning}
\subsubsection{Construction of multi-layered road-graph}
One of the most efficient methods of autonomous driving is to use a 2-D semantic road map, which $M_{2D}^W$ can be utilized for behavior planning and identifying environmental recognition failures.
Therefore, we can construct a 2-D semantic road map $M_{2D}^W$ by accumulating a road-marker feature $\mathcal{F}_{road, rgb}^B$ that is detected with a deep-learning-based lane detector \cite{wang2018lanenet}. 
At the same time, the result of the bird’s eye view feature point $\mathcal{F}_{road, rgb}^W$ is accumulated on the global coordinates according to $\hat{\mathbf{x}}_{t}$. 
After constructing a primary semantic map, we refine $\mathcal{F}_{road, rgb}^W$ to build $M_{2D}^W$.
As an equivalent to $\mathcal{F}_{road, rgb}^W$, point-cloud based information $\mathcal{F}_{road,point}^W$ is accumulated, where ${F}_{road,point}^B$ is generated using point reflectivity and a high-pass filtering mask \cite{pardo2006downscaling}.
Therefore, we convert unordered raw point-cloud data to ordered point-cloud data $\{{P}_i| i = 1,...,n\}$, where each point $P_i$ is a vector of $(x,y,z,r)$, to compute $\mathcal{F}_{road,point}^B$ as, 
\begin{equation}
\begin{aligned}
    \mathcal{F}_{road,point}^B = {P}(r_{i=1:n}) * H_{sharp},
\end{aligned}
\label{eq_road_feature_point}
\end{equation}
where $*$ is a convolution operation, $H_{sharp}$ is a high-pass filtering(sharpening) mask, and $r$ is reflectivity. \\
As a result, we can define the semantic map $M_{2D}^W$ as being composed of segmented links $\mathbf{E}_{1:n}= ({E}_{1,j1,k1},\dots, {E}_{n,jn,kn})$, segmented lanes $\mathbf{L}_{1:m}= ({L}_{1,l,r},\dots, {L}_{m,l,r})$ and nodes $\mathbf{N}_{1:l,i} = ({N}_{1,i},\dots, {N}_{l,i}) \in \mathbf{E}_{i}$ where links are accumulated vehicle poses $\hat{\mathbf{x}}_{t}$ and lanes are detected road-marker features $\{\mathcal{F}_{road,rgb}^W, \mathcal{F}_{road,point}^W\} \in {L}_{m,l,r}$.
Each individual link ${E}_{i,ji,ki} \ni \{E_{ji}, E_{ki}\}$ and segmented lane ${L}_{m,l,r}$ has attributes such as hash-ID, position, and road type.
Moreover, we refine the road map using a handcrafted method and include additional attributes for parallel right $E_{ji}$ and left $E_{ki}$ links, which can be utilized for the behavior planning and path planning.
Therefore, we constructed our own point-cloud map and road graph, to ensure the capability of managing the changes in the real-world environment.
\begin{figure}[t!]
\centering
\includegraphics[width=1.0\columnwidth]{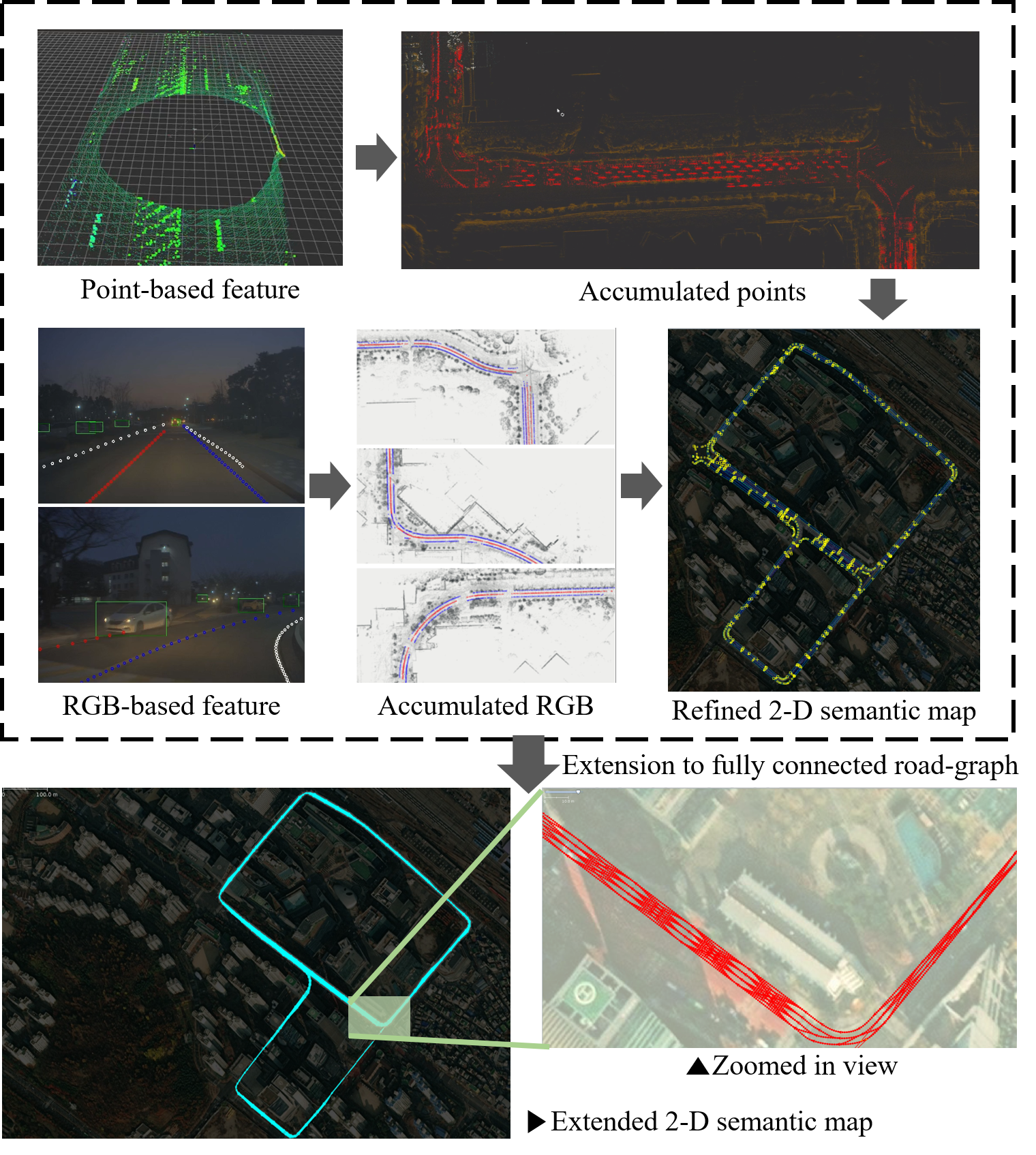}
\caption{The process of building a baseline and an extended 2-D semantic road graph, which is built using both RGB-based and point-cloud-based features. Subsequently, to extend the 2-D semantic road map, a fully connected road graph is generated.}
\label{fig:semantic_map_build}
\end{figure}

\subsubsection{Graph-searching based route planning}
Given the estimated position $\hat{\mathbf{x}}$, the vehicle is expected to drive along a route to reach the target destination. 
Let the optimal route be defined as $\mathbf{p}^{g*}_{1:n} = \{p_1^g, \dots, p_n^g\}$, with size $n$ of poses. 
Specifically, the cost of each segmented link is $E_{i, ji, ki}$ as $f^g(E_i)$. 
We define $P_g(E_i) = \sum_{i = 0}^{n}f^g(E_i)$ as the accumulated cost of the consecutive link to the destination node $N_{G} = \{x_g, y_g\}$.
Then, we can define the route planning problem as
\begin{equation}
\begin{aligned}
    \mathbf{p}^{g*} = \arg\min_{\forall{i}}{P_g(E_{i, ji, ki} | N_{S} )},
\end{aligned}
\label{eq_route_planning}
\end{equation}
where $N_{S}$ is the start node.
Assuming that road graph $M_{2D}^W = \{\mathbf{N}_{i,1:n}, \mathbf{E}_{1:n}\}$ is built to connect all the previous and subsequent $N_{j,i}$ and $E_{i, ji, ki}$, 
the planning problem $\mathbf{p}^{g*}$ can be defined as searching for a continuous route from the start node $N_{S}$ to the target node $N_{G}$, based on ${G}$.
Because the graph-searching based algorithm finds routes on the grid \cite{duchovn2014path}, we have adopted the analogous concept for the road-graph.
When $N_{S}$ is defined from $\hat{\mathbf{x}}$, we maximize $P_g(\cdot)$ by propagating the graph to arrive at the destination.
When the graph-searching reaches the destination, an optimal route is obtained on the visited node using a depth-first searching algorithm \cite{tarjan1972depth}.
If a route is not obtained on the single consecutive road-graph, $\mathbf{p}^{g*}_{1:n}$ is computed based on the propagating parallel lane because our graph can find parallel links, which are represented as $E_{ji}$ and $E_{ki}$ for the left and right parallel links from $E_{i}$, respectively.

\subsubsection{Extended road-graph}
The extended 2-D semantic road graph $M_{ext}^W = \{N_{i,1:m}, E_{1:m}, C_{i: 1:m}, \kappa_{i, 1:m}\}$ is defined by further enriching connectivity $C_{i: 1:m}$ and curvature $\kappa_{i, 1:m}$ and interpolating between multiple lanes.
Here, we adopt the ideas in \cite{stahl2019multilayer} to precompute the state lattice offline, which is defined in the Frenét space along a $\mathbf{p}^{g*}_{1:n}$, where the Frenét frame is defined as the coordinate system spanned by the tangential and normal vectors at any point of the reference line.
The state lattice is defined as a discrete function $[x(s), y(s),\theta(s),\kappa(s)]$ along the arc length $s$, also called the station. 
Here, the refined $M_{ext}^W$ is formed by lattice layers distributed along the station $s$ connecting interpolated splined points as depicted in Fig. \ref{fig:semantic_map_build}.
We utilize $M_{ext}^W$ for trajectory prediction of objects. 
Furthermore, we present the road-graph-searching-based motion planning algorithm using $M_{ext}^W$ in Section \ref{sec:motion_planning}. 

\subsection{Motion planning}
\label{sec:motion_planning}
The task of the motion planning module is to find a collision-free and dynamically feasible path to arrive at a specific goal configuration. There are several major methods in the motion planning field: grid-based, sampling-based, and motion-primitive-based planning. 
The graph-based motion planners, such as the A* search algorithm \cite{a_star, a_star_variation, hybrid_a_star, hybrid_a_star_parking, paden2016survey}, find the shortest path based on an admissible heuristic cost function \cite{a_star, a_star_variation}. 
The planner searches collision-free nodes with a grid-map-based graph representation. 
Variations of the A* algorithm consider the non-holonomic constraint of autonomous vehicles for urban environment applications \cite{hybrid_a_star, hybrid_a_star_parking}. 
To extend the search space to continuous coordinates, hybrid states containing the discretized cell information and continuous 3-D states $(x, y, \theta)$ were configured. 
The hybrid representation guarantees the algorithm will search kinematically feasible trajectories for urban driving \cite{paden2016survey}. 
The sampling-based methods explore the configuration space using probabilistic node sampling to find a feasible path. A tree of collision-free paths is incrementally extended using random or heuristic \emph{steering} and \emph{collision test} functions \cite{rrt_original, rrt_darpa}. 
With additional routines, such as \emph{rewiring}, these methods trim the redundant nodes in the tree to obtain a minimum-cost path to a goal configuration \cite{rrt_star_original, informed_rrt_star}. 
Because of the probabilistic completeness, most of the research focuses on exploring unstructured environments \cite{rrt_exploration_mobile, rrt_exploration_drone} or planning for high-dimensional systems \cite{kinodynamic_rrt_star, sampling_with_constraints}.

\subsubsection{Road-graph searching based macro-motion planning}
Motion planning based on the graph-searching algorithm \cite{a_star, a_star_variation} is considered a milestone in robotics for the solution of obstacle avoidance problems.
However, a grid-based graph-searching algorithm is not adaptable for autonomous driving because it does not consider traffic laws.
We propose an extended 2-D semantic road graph $M_{ext}^W$ that enriches connectivity and interpolates between multiple lanes. 
In this study, we divide the motion planning algorithm into macro and micro scale, and propose a road-graph based searching algorithm for macro-motion planning.
The goal of the macro-motion planning algorithm is to find an optimal trajectory $\mathbf{p}^{macro*}_{t, 1:n}$ considering obstacles, lane changes and vehicle dynamics.
We design the heuristic cost $g^{macro}_{i,1:m}$ $\in$ ${N}_{i,1:n}$ as   
\begin{equation}
\begin{aligned}
    g^{macro}_{i, 1:m} = 
    & k_{\mathcal{O}}\parallel\mathcal{O}_{i} - {N}_{i, 1:m}\parallel + k_{\kappa}{\kappa}_{i, 1:m} + k_{trans}\delta^{macro}_{i} + \\
    & k_{route}\parallel\mathbf{p}^{g*}_{1:n} - {N}_{i, 1:m}\parallel,
\end{aligned}
\label{eq_macro_path}
\end{equation}
considering the distance from the closest node ${N}_{i, 1:m}$ to an obstacle $\mathcal{O}_{i}$, road-curvature $\kappa_{i}$, transient path $\delta^{macro}_{i}$, distance from the closest node ${N}_{i, 1:m}$ to planned route $\mathbf{p}^{g*}_{1:n}$, and weight for each cost $k_{\mathcal{O}}$, $k_{\kappa}$, $k_{trans}$, $k_{route}$, respectively.
In addition, $\delta^{macro}_{i}$ is considered not to change trajectory dynamically as
\begin{equation}
\begin{aligned}
    \delta^{macro}_{i} = \parallel \mathbf{p}^{macro*}_{t-1, 1:n} - {N}_{i, 1:m}\parallel,
\end{aligned}
\label{eq_macro_transient}
\end{equation}
where $\mathbf{p}^{macro*}_{t-1, 1:n}$ is the optimal trajectory of the previous step, to ensure $\mathbf{p}^{macro*}_{t, 1:n}$ not to be a large change.
Therefore, a macro-motion planning algorithm can be calculated considering both travel distance and designed heuristic cost as  
\begin{equation}
\begin{aligned}
    \mathbf{p}^{macro*}_{t, 1:n} = arg\min_{\forall{i}}{(f^{macro}_{i, 1:m} + g^{macro}_{i, 1:m})},
\end{aligned}
\label{eq_macro_planning}
\end{equation}
where travel distance cost $f^{macro}_{i, 1:m}$ is obtained through the accumulating distance of nodes.

\subsubsection{Motion primitives-based micro-motion planning}
Although the macro-motion trajectory $\mathbf{p}^{macro*}_{t, 1:n}$ is already computed, considering obstacles and the road-curvature model, the vehicle cannot possibly avoid obstacles within a safe margin.
An accident can happen when the vehicle cannot accurately follow the trajectory due to vehicle dynamic characteristics, such as tire slip or vehicle control performance. 
Therefore, we utilize a motion primitive-based micromotion planning algorithm that has been widely adopted in recent work \cite{werling2010optimal, zhu2020trajectory, zheng2020bezier}. \\
We define a set of positions with size $k$ $\mathbf{p}^{micro}_{1:k, j} = \{p_{1,j}^{micro}, \dots, p_{k,j}^{micro}\}$, where each position is defined as $p_{i,j}^{micro} = \{x_{i,j}^{micro}, y_{i,j}^{micro}\}$.
Therefore, a set of motion primitives can be defined as $\mathbf{P}^{micro}_{1:j} = [\mathbf{p}^{micro}_{1:k, 1}, \dots, \mathbf{p}^{micro}_{1:k, j}]$, where $j$ is the number of motion primitives.
We generate a motion primitive $\mathbf{p}^{micro}_{1:k, j}$ derived from polynomial and macro planning $\mathbf{p}^{macro*}_{t, 1:n}$.
First, we define $\mathbf{x}_s = \{x_s, \Dot{x}_s, \Ddot{x}_s, y_s, \Dot{y}_s, \Ddot{y}_s\}$ and $\mathbf{x}_f = \{x_f, \Dot{x}_f, \Ddot{x}_f, y_f, \Dot{y}_f, \Ddot{y}_f\}$ as the start and final states, respectively, and the time interval between the start and final states as ${\Delta}t_f = t_f - t_s$.
Then, $\mathbf{p}^{micro}_{1:k, j}$ is generated by calculating each $p_{i,j}^{micro} = \{x_{i,j}^{micro}, y_{i,j}^{micro}\}$, as follows: 
\begin{equation}
    \begin{aligned}
    & x_{i,j}^{micro}(t) = a_0 + a_{1}t + a_{2}t^2+ a_{3}t^3+ a_{4}t^4 + a_{5}t^5, \\
    & y_{i,j}^{micro}(t) = b_0 + b_{1}t + b_{2}t^2+ b_{3}t^3+ b_{4}t^4 + b_{5}t^5,
    \end{aligned}
\end{equation}
where the polynomial coefficients for $x_{i,j}^{micro}$ and $y_{i,j}^{micro}$ are denoted by $[a_{0},a_{1},a_{2},a_{3},a_{4},a_{5}]^{T}$ and $[b_{0},b_{1},b_{2},b_{3},b_{4},b_{5}]^{T}$, respectively.
Here, we can update the initial state of $\mathbf{x}_s$ using the wheel encoder and IMU. 
Furthermore, we selected a series of target states to be determined using $w_{road}$, and the target speed and acceleration at $\mathbf{x}_f$. 
Our motion primitive generator can be expressed as follows:
\begin{equation}
\begin{aligned}
    \begin{bmatrix} 
       a_{0} \\ a_{1} \\ a_{2} \\ a_{3} \\ a_{4} \\ a_{5}
    \end{bmatrix} = 
    \begin{bmatrix} 
        1 & t_{s} & t_{s}^2 & t_{s}^3 & t_{s}^4 & t_{s}^5\\
        0 & 1 & 2t_{s} & 3t_{s}^2 & 4t_{s}^3 & 5t_{s}^4 \\
        0 & 0 & 2 & 6t_{s} & 12t_{s}^2 & 20t_{s}^3 \\
        1 & t_{f} & t_{f}^2 & t_{f}^3 & t_{f}^4 & t_{f}^5\\
        0 & 1 & 2t_{f} & 3t_{f}^2 & 4t_{f}^3 & 5t_{f}^4 \\
        0 & 0 & 2 & 6t_{f} & 12t_{f}^2 & 20t_{f}^3
    \end{bmatrix}^{-1}    
    \begin{bmatrix} 
    x_s\\ \Dot{x}_s \\ \Ddot{x}_s\\ x_f\\ \Dot{x}_f \\ \Ddot{x}_f 
    \end{bmatrix}, 
\end{aligned}
\label{eq:poly_x}
\end{equation}
\begin{equation}
\begin{aligned}
    \begin{bmatrix} 
       b_{0} \\ b_{1} \\ b_{2} \\ b_{3} \\ b_{4} \\ b_{5}
    \end{bmatrix} = 
    \begin{bmatrix} 
        1 & t_{s} & t_{s}^2 & t_{s}^3 & t_{s}^4 & t_{s}^5\\
        0 & 1 & 2t_{s} & 3t_{s}^2 & 4t_{s}^3 & 5t_{s}^4 \\
        0 & 0 & 2 & 6t_{s} & 12t_{s}^2 & 20t_{s}^3 \\
        1 & t_{f} & t_{f}^2 & t_{f}^3 & t_{f}^4 & t_{f}^5\\
        0 & 1 & 2t_{f} & 3t_{f}^2 & 4t_{f}^3 & 5t_{f}^4 \\
        0 & 0 & 2 & 6t_{f} & 12t_{f}^2 & 20t_{f}^3
    \end{bmatrix}^{-1}    
    \begin{bmatrix} 
    y_s\\ \Dot{y}_s \\ \Ddot{y}_s\\ y_f\\ \Dot{y}_f \\ \Ddot{y}_f 
    \end{bmatrix}.
\end{aligned}
\label{eq:poly_y}
\end{equation}
Subsequently, we convert the generated $\mathbf{p}^{micro}_{1:n}$ consisting of $p_{i,j}^{micro} = \{x_{i,j}^{micro}, y_{i,j}^{micro}\}$ into curvilinear coordinates.
Therefore, a single motion primitive, i.e., the set of positions $\mathbf{p}^{micro}_{1:k, j} = \{p_{1,j}^{micro}, \dots, p_{k,j}^{micro}\}$, is modeled using an arc-length-based cubic spline, such that
\begin{equation}
    \begin{aligned}
    \mathbf{s}^{micro}_{n, j} &= 
    \sum_{i = 0}^{n - 1}{\sqrt{ (x_{i+1,j}^{micro} -  x_{i,j}^{micro})^2 +  
    (y_{i+1,j}^{micro} -  y_{i,j}^{micro})^2}}, \\
    &= \sum_{i = 0}^{n - 1}{{s}^{micro}_{i, j}}. 
    \end{aligned}
    \label{eq:cuv_coord}
\end{equation}
We define the motion primitives converted to curvilinear coordinates as $\mathbf{P}^{s}_{1:j} = [\mathbf{p}^{s}_{1:k, 1}, \dots, \mathbf{p}^{s}_{1:k, j}]$. 
We also label a converted single primitive $\mathbf{p}^{s}_{1:k, j} = \{p_{1,j}^s, \dots, p_{k,j}^s\}$.
Our final motion primitive can be computed as
\begin{equation}
    \begin{aligned}
        x_{i,j}^s =& a_{x}({s}^{micro}_{i, j} - s^g)^3 + b_{x}({s}^{micro}_{i, j}- s^g)^2 + \\                 & c_{x}({s}^{micro}_{i, j}- s^g) + d_x, \\
    \end{aligned}
\end{equation}
\begin{equation}
    \begin{aligned}
        y_{i,j}^s = & a_{y}({s}^{micro}_{i, j} - s^g)^3 + b_{y}({s}^{micro}_{i, j}- s^g)^2+ \\
                    & c_{y}({s}^{micro}_{i, j}- s^g) + d_y, \\
    \end{aligned}
\end{equation}
where the arc length $s^g$ is the cumulative distance sum of referential path $\mathbf{p}^{macro*}_{1:n}$. 
In addition, coefficients $a_{x,y}, b_{x,y}, c_{x,y}$, and $d_{x,y}$ of the cubic spline can be calculated using the boundary conditions of the first and second derivatives \cite{conte2017elementary}. 
We propose cost function $g^{micro}(\mathbf{p}^{s}_{1:k, i})$ to obtain a micro optimal trajectory $\mathbf{P}^{micro*}_{t, 1:n}$ considering obstacles, curvature and vehicle dynamics as
\begin{equation}
    \begin{aligned}
    g^{micro}(\mathbf{p}^{s}_{1:k, i}) = {w_{\mathcal{O}}}{\mathcal{O}_{1:k}} + {w_\kappa}{\lVert{\kappa_{1:k}}\rVert} + {w_{trans}}(\delta^{micro}_{i}), 
    \end{aligned}
    \label{eq:micro_planning}
\end{equation}
where $\mathcal{O}_{1:k}$ is the distance to obstacles corresponding to the closed position $\{x_{1:k}, y_{1:k}\}$,
the second term represents the sum of the curvature of the primitive; 
the third term is the final transient state compared with the previous optimal primitive,
and the weight for each cost $w_{\mathcal{O}}$, $w_{\kappa}$, and $w_{trans}$, respectively.
In addition, $\delta^{micro}_{i}$ is considered not to change trajectory dynamically as
\begin{equation}
\begin{aligned}
    \delta^{micro}_{i} = \parallel \mathbf{P}^{micro*}_{t-1, 1:n} - \mathbf{p}^{s}_{1:k, i}\parallel,
\end{aligned}
\label{eq_micro_transient}
\end{equation}
where $\mathbf{P}^{micro*}_{t-1, 1:n}$ is the micro optimal trajectory of the previous step, to ensure $\mathbf{P}^{micro*}_{t, 1:n}$ not to be a large transient.
Therefore, micro-motion planning algorithm can be calculated considering both travel distance and designed heuristic cost as  
\begin{equation}
\begin{aligned}
    \mathbf{p}^{micro*}_{t, 1:n} = arg\min_{\forall{i}}{(f^{micro}_{i, 1:m} + g^{micro}_{i, 1:m})},
\end{aligned}
\label{eq_micro_planning}
\end{equation}
where travel distance cost $f^{micro}_{i, 1:m}$ is obtained by the accumulating distance of nodes.

\subsection{Behavior planning}
\label{sec:behavior}
We proposed a route planning algorithm to calculate the minimum-transversal-distance route on the road-graph in \ref{sec:route_planning}.
While driving to the destination in a congested urban environment, there are a few scenarios for overtaking other vehicles. 
Specifically, a complex behavior planning strategy is demanded for safe overtaking. The planning algorithm must be able to execute various lateral maneuvers considering the location, intentions of surrounding vehicles, and signals from the traffic infrastructure. 
Furthermore, because a naive conservative-spacing algorithm may prohibit the reduction of headway and discourage overtaking maneuvers, the ego vehicle needs to infer efficient longitudinal velocity according to the distances from not only the front vehicle but also the vehicle on the overtaking side.\\
To tackle those above-mentioned challenges, we designed a task-specific path selection (TSPS) algorithm that can derive a feasible high-level lateral maneuver considering the surrounding vehicles, traffic environment, and state and intent of the ego vehicle. 
Furthermore, we designed a geometry-aware velocity planning (GVP) algorithm that can consider the geometric relationship between the ego and surrounding vehicles for safe overtaking.

\subsubsection{Task-specific path selection}
We present a TSPS module to consider multiple options and determine optimal high-level lateral maneuvering for various scenarios.
To simplify this complex behavior planning problem, we divide lateral maneuver into three path options : global optimal route, overtaking/avoidance trajectory, and ego lane.
More precisely, ego lane is computed from the pre-built road-graph.
Let our neighbor road-graph set $\mathbf{p}^{n.b}_{t, 1:n}$ be defined as
\begin{equation}
\begin{aligned}
    \mathbf{p}^{n.b}_{t, 1:n} = \{N_{i,j}, E_{j} | d_{j}^{n.b}, \hat{\mathbf{x}}_t \}_{j \in [1:n]},
\end{aligned}
\label{eq_neighbor}
\end{equation}
where $d_{j}^{n.b}$ is a Euclidean distance from $\hat{\mathbf{x}}_t$ to the road-graph set. 
Subsequently, we can extract the closest lane to the vehicle in the road-graph set as 
\begin{equation}
\begin{aligned}
    \mathbf{p}^{ego}_{t, j} = arg\min_{\forall{d_j^{n.b}}}(\mathbf{p}^{n.b}_{t, 1:n}).
\end{aligned}
\label{eq_ego_lane}
\end{equation}
As a result, we implement the TSPS algorithm to derive a feasible path from a set of path models from the original route, motion-planning path, and closest lane from the road graph, respectively, $\mathbf{p}^{g*}$, $\mathbf{p}^{micro*}_{t, 1:n}$, and $\mathbf{p}^{ego}_{t, j}$ as
\begin{equation}
\begin{aligned}
    \mathbf{p}_{t}^{tsps}
    = 
    \begin{cases}
    &  \mathbf{p}^{g*} \\
    &  \mathbf{p}^{micro*}_{t, 1:n} \\
    &  \mathbf{p}^{ego}_{t, j} \\
    \end{cases}
\end{aligned}
\label{eq_tsps}
\end{equation}
The TSPS algorithm follows the hierarchical process shown in Fig. 7.
First, if the original global route is feasible, the TSPS derives this route as the final path without additional operation.
If not, the TSPS then checks whether the ego vehicle is stuck. 
Except in the case of a red traffic signal, the algorithm accumulates a stuck counter at every moment of zero velocity status and decides whether the vehicle is stuck according to the value of the counter.
Because a long stuck status degrades the overall progress of the ego vehicle, after a few stuck steps, the TSPS derives a micro-motion path to induce the ego vehicle to escape from the stuck situation.
If the ego vehicle is in normal status, the TSPS examines the viability of the local path (micro path). If the path is not feasible, it derives the original route for the ego vehicle to keep its lane without changing lanes or overtaking.
If the vehicle is close to a traffic signal, the TSPS considers the following driving scenarios: turn-left, turn-right, or go-straight. 
Because the ego vehicle needs to follow the direction of the route exactly, the algorithm outputs the original route as the final path when the scenario is turn-left or turn-right for the vehicle not to deviate from the original route. In the go-straight scenario, the algorithm returns the micro-local path, as the dependency on the original route direction is not prominent.
In the aforementioned process, even if the local path is feasible, following the local path and performing immediate overtaking is risky. 
The safety of the side area must be considered before executing lateral maneuvers through the micro-local path..
Therefore, if there is a surrounding vehicle in the side area, the TSPS outputs the closest neighboring path instead of the local path for the ego vehicle to keep its lane, while maintaining a safe distance. Consequently, the ego vehicle performs safe lateral maneuvers only when the safe-overtaking condition is satisfied.

\begin{figure}[t!]
\centering
\includegraphics[width=0.95\columnwidth]{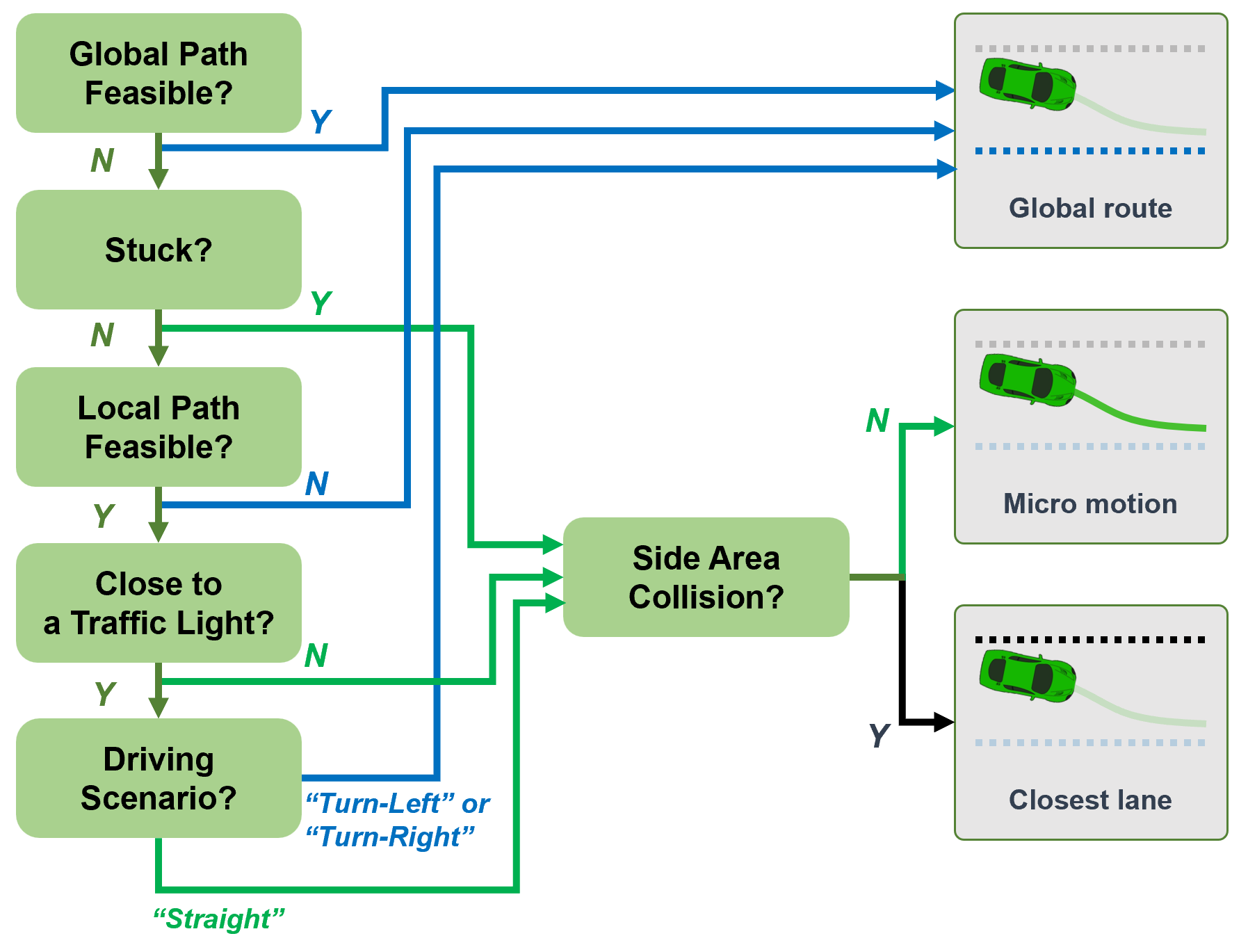}
\caption{Mechanism of the decision-tree-based path selection algorithm.}
\label{fig:decision_tree}
\end{figure}

\subsubsection{Geometry-aware velocity planning}
For safe overtaking in urban environments, the ego vehicle must plan progressive longitudinal maneuvers, while maintaining safe headway from the vehicles on both the front and overtaking sides. 
However, a naive velocity-planning strategy, such as adaptive cruise control, generates velocity commands considering the forward headway only, which may not allow reducing the distance from the front vehicle and may discourage overtaking maneuvers. 
Moreover, because the ego vehicle’s overtaking switches the target of the front vehicle, it is necessary to consider the geometric relationship between the ego and surrounding vehicles during the overtaking transition. 
Therefore, we designed the GVP algorithm to consider the dynamic-geometric relationship between the ego and surrounding vehicles for safe overtaking scenarios. 
We implemented the algorithm with design factors that can 1) operate in both general-lane-following and overtaking scenarios with a single algorithm and 2) alter the aggressiveness of overtaking through simplified parameters.

The GVP calculates two velocity commands, $v_{T}$ and $v_{E}$, which are computed using an adaptive cruise control algorithm considering the vehicle on the target lane $\mathbf{p}^{target}_{t}$ and ego lane $\mathbf{p}^{ego}_{t, j}$ respectively. Those velocity commands are used to derive new velocity plans $v^{micro}$ and, $v^{cog}$ considering the road geometry during overtaking as
\begin{equation}
    \label{eq:wacc_1}
    \begin{aligned}
    & v^{micro} = (1 - \frac{{d^T}_t}{{d^T}_t + {d^E}_t}) v_T + (1 - \frac{{d^E}_t}{{d^T}_t + {d^E}_t}) v_E \\
    & v^{cog} = (1 - \frac{{d^T}_e}{{d^T}_{e} + {d^E}_e}) v_T + (1 - \frac{{d^E}_e}{{d^T}_{e} + {d^E}_e}) v_E
    \end{aligned}
\end{equation}
where $d^T_t$ and, $d^E_t$ are the distances from the terminal point of $\mathbf{p}^{micro*}_{t, 1:n}$ to $\mathbf{p}^{target}_{t}, \mathbf{p}^{ego}_{t, j}$ respectively, and $d^T_{e}$ and $d^E_{e}$ are the distances from the center of gravity of the ego vehicle to $\mathbf{p}^{target}_{t}$ and, $\mathbf{p}^{ego}_{t, j}$, respectively.

The final velocity command of GVP is then computed as
\begin{equation}
    \label{eq:wacc_2}
    \begin{aligned}
    & v^{ref} = \tau v^{micro} + (1 - \tau) v^{cog}
    \end{aligned}
\end{equation}
where $\tau$ is the aggressiveness factor. 
Consequently, vehicle operators could set a hyper-parameter $\tau$ before autonomous driving begins.

\begin{figure}[t!]
\centering
\includegraphics[width=0.7\columnwidth]{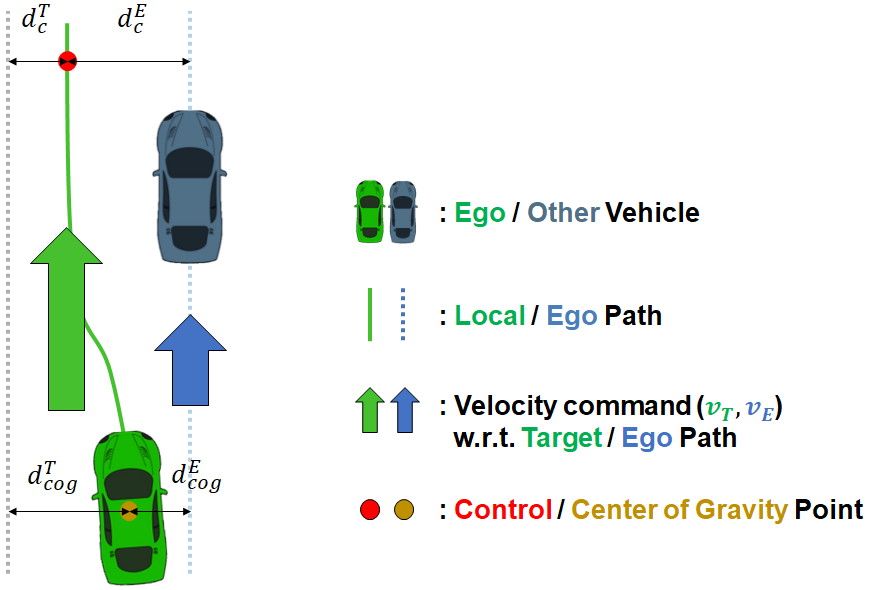}
\caption{Schematic of the proposed geometry-aware velocity-planning algorithm in an overtaking scenario.}
\label{fig:WACC}
\end{figure}

\section{Traffic Analysis}
\label{sec:traffic}
In Section \ref{sec:behavior}, we proposed a driving model using TSPS and GVP algorithms, which can determine the aggressiveness of the driving model for overtaking and lane changes.
Additionally, we designed a behavior planning algorithm that will obey traffic laws at the critical penalty area. Owing to these traffic constraints, most vehicles will drive analogously and generate traffic congestion. 
Therefore, we have expanded our study to include traffic analysis for multiple autonomous vehicles in traffic-congested situations. \\
In recent years, there have been numerous studies on cooperative traffic systems utilizing V2X communication, which includes vehicle-to-vehicle(V2V) and V2I. 
Without cooperative driving using communication, autonomous vehicles can have conflicting optimal conditions.
Especially, in an autonomous driving competition, each team can assume that every other team will drive competitively, rather than cooperatively, for the race.
Our concern was similar to this conflict that transversal-time-minimizing planning can be analogous between autonomous vehicles considering minimum distance, curvature, or transversal time. \\
In this section, we present a methodology for traffic analysis to determine the conflict of autonomous vehicles’ optimality. 
Specifically, to measure the similarity of spatial distribution, we measure a Kullback–Leibler divergence (KLD) estimation for spatial data $P(\mathbf{x|z})$ and $Q(\mathbf{x|z})$, with a finite k. 
In \cite{perez2008kullback}, the author proved the estimation of the divergence for vectorial data using the k-th nearest-neighbor density estimate as
\begin{equation}
\begin{aligned}
    D_{\mathbb{KL}}^{spatial}(P \parallel Q) = & \sum_{\mathbf{x}\in\mathcal{X}} P_{k}(\mathbf{x|z}) \log\left(\frac{P_{k}(\mathbf{x|z})}{Q_{k}(\mathbf{x|z})}\right),
\end{aligned}
\label{eq_vectorial_kl_divergence}
\end{equation}
where $\mathbf{z}$ is the reference racing line or road-graph and $P(\mathbf{x|z})$ can be considered the conditional distribution of the spatial data. 
Assuming the given spatial data $P(\mathbf{x|z})$ and $Q(\mathbf{x|z})$ have $m$ and $n$ samples of 2-D data, respectively, we can define the KLD for the spatial distribution as
\begin{equation}
\begin{aligned}
    D_{\mathbb{KL}}^{spatial}(P \parallel Q) = & \sum_{\mathbf{x}\in\mathcal{X}} P_{k}(\mathbf{x|z}) \log\left(\frac{r_{k}(\mathbf{x|z})}{s_{k}(\mathbf{x|z})}\right) \\
                                               & + \log\left (\frac{m}{n-1}\right),
\end{aligned}
\label{eq_spatial_kl_divergence}
\end{equation}
where
\begin{equation}
\begin{aligned}
    &P_{k}(\mathbf{x|z}) = \frac{k}{(n-1)}\frac{\Gamma(d/2 + 1)}{\pi^{d/2}r_{k}(\mathbf{x}^d)}, \\
    &Q_{k}(\mathbf{x|z}) = \frac{k}{m}\frac{\Gamma(d/2+1)}{\pi^{d/2}s_{k}(\mathbf{x})^d}.
\end{aligned}
\label{eq_spatial_kl_divergence_r_and_s}
\end{equation}
In Eq. \eqref{eq_spatial_kl_divergence_r_and_s}, $r_k(\mathbf{x})$ and $s_k(\mathbf{x})$ are the Euclidean distances to the \textit{k}-th nearest neighbor of $\mathbf{x} \in  \mathcal{X}$.
In addition, $\pi^{d/2} / \Gamma(d/2+1)$ is the volume of the unit-ball in $\mathbb{R}^d$. \\
Furthermore, we analyze the small-scale data with the Euclidean distance error between the two autonomous vehicles data $\mathcal{X}, \mathcal{Y}$ to understand the mean error.
We compute the mean error value as
\begin{equation}
\begin{aligned}
    \Bar{d}_{\mathcal{X}, \mathcal{Y}} = & \frac{1}{n} \sum_{\mathbf{x}\in\mathcal{X}, \mathbf{y}\in\mathcal{Y}} (\mathbf{x} -  r_k(\mathbf{y})),
\end{aligned}
\label{eq_mean_distance_error}
\end{equation}
where $n$ is the size of the data $\mathcal{X}$, and $r_k(\mathbf{x})$ is the Euclidean distances to the \textit{k}-th nearest-neighbour of $\mathbf{x} \in  \mathcal{X}$. \\
As a result, we can utilize Eqs. \eqref{eq_spatial_kl_divergence} and \eqref{eq_mean_distance_error} to understand the spatial data of various autonomous vehicles in the scope of traffic analysis. 

\section{Results}
\label{sec:experiments}
\subsection{Test environment}
\subsubsection{Sangam dataset}
In this study, several experiments were conducted using the Sangam real-world dataset that contains 128-channel LiDAR points, front and rear RGB camera images, a single IMU, and vehicle-state data. 
We also annotate data from vehicles that can be utilized for object detection and tracking. 
In the real world, we evaluated the proposed autonomous stack of system modules in a mixed traffic environment, where both autonomous and human-driving vehicles were deployed.

\subsubsection{Simulation dataset}
We also present a simulation environment that enables us to evaluate our proposed system in various scenarios. 
When evaluating algorithms for motion and behavior planning, it is difficult to repeatedly perform the same experiment through field testing. To end this, we utilized the IPG CarMaker simulator distributed by Hyundai Motor Company. 
In the simulation, the Sangam area, where field tests were conducted, is simulated, allowing us to implement virtual test scenarios for the autonomous vehicles in the application areas, with high-resolution 3-D visualization photorealistic quality.

\subsection{Resilient localization}
We evaluated our proposed resilient localization system, which does not depend on GPS, in the urban city area. 
Our evaluation of localization can be divided into two components: registration algorithm selection and performance of the resilient system. In the pioneering studies \cite{gicp, koide2021voxelized, sommer2020openmp, rusu2009fast,salti2014shot, chetverikov2002trimmed}, the authors proposed a series of registration algorithms that can measure corresponding points between 3-D input and target data.
Thanks to these studies, we can implement a non-GPS-based localization system utilizing a 3-D pre-built point-cloud map.
We considered two main real-time registration algorithms that are implemented using multi-threading accelerated computing, named Voxelized-GICP \cite{koide2021voxelized} and NDT-OMP \cite{biber2003normal, koide2019portable}.
\begin{table*}[ht]
\caption{Parameters for registration selection considering success/failure, computing stress, and range. 
We conducted a parametric study by altering a couple of parameters.}
\label{tab:localize_table}
\centering
\resizebox{2\columnwidth}{!}
{\begin{tabular}{c|c|c|c|c|c|c|c|c|c||c}
\hline
Methods & 
\begin{tabular}[c]{@{}c@{}}Num. of \\ input cloud channel \end{tabular} & 
Sensing radius [m] &  
\begin{tabular}[c]{@{}c@{}}Sliding-window map \\ radius [m] \end{tabular}& 
Voxel size [m] & 
\begin{tabular}[c]{@{}c@{}}Matching error \\ threshold \end{tabular} & 
Maximum iteration & 
\begin{tabular}[c]{@{}c@{}}Transformation \\ epsilon [m] \end{tabular} & 
Num. of thread & 
\begin{tabular}[c]{@{}c@{}}Maximum \\ corresponding distance [m] \end{tabular}& 
Outcome \\
\hline
NDT-OMP & 128 & 70 & 70 & 0.5 & 0.2 & 32 & 0.01 & - & - & Fail\\
NDT-OMP & 64(upper)  & 70 & 70 & 0.5 & 0.2 & 64 & 0.01 & - & - & Fail\\
NDT-OMP & 64(upper) & 100 & 100 & 0.5 & 0.2 & 128 & 0.01 & - & - & Fail\\
NDT-OMP & 64(upper) & 80 & 80 & 0.5 & 0.2 & 64 & 0.03 & - & - & Success\\
\hline
Voxelized-GICP & 96(upper) & 100 & 100 & 1.0 & 0.5 & - & - & 2 & 1.0 &Fail\\
Voxelized-GICP & 96(upper) & 100 & 100 & 1.0 & 0.5 & - & - & 10 & 1.0 &Fail\\
Voxelized-GICP & 96(upper) & 100 & 100 & 1.0 & 0.5 & - & - & 6 & 1.0 &Fail\\
Voxelized-GICP & 128 & 100 & 100 & 1.0 & 0.5 & - & - & 6 & 3.0 &Success\\
\hline
\end{tabular}}
\end{table*} \\
As shown in Table \ref{tab:localize_table}, parameters of registration affect the performance of the scan matching-based localization algorithm.
To identify the impact of each parameter, we conducted a parametric study by altering the input cloud channel, sensing radius, sliding-window map radius, voxel size, and matching error threshold. 
We focused on determining the desired output that can cover the urban map scale, computing time, and performance at high speed. 
While selecting the registration algorithm, we implemented our proposed sliding-window-based registration algorithm because using the entire map is too slow to run even a few steps of registration.
However, only the registration-based scan-matching algorithm has a limitation in a large-scale environment. 
Therefore, we implemented robust navigation using LiDAR odometry for registration algorithm evaluation, and our proposed method could determine the dynamic movement of the vehicle. 
After evaluating a series of registrations with various parameters, we elected to use Voxelized GICP with a coarse voxel size of 1.0 $m$, which was verified up to a driving speed of 60 $kph$ in the urban area.
\begin{figure}[!t]
    \centering
    \subfigure[]{
        \includegraphics[width=0.9\columnwidth]{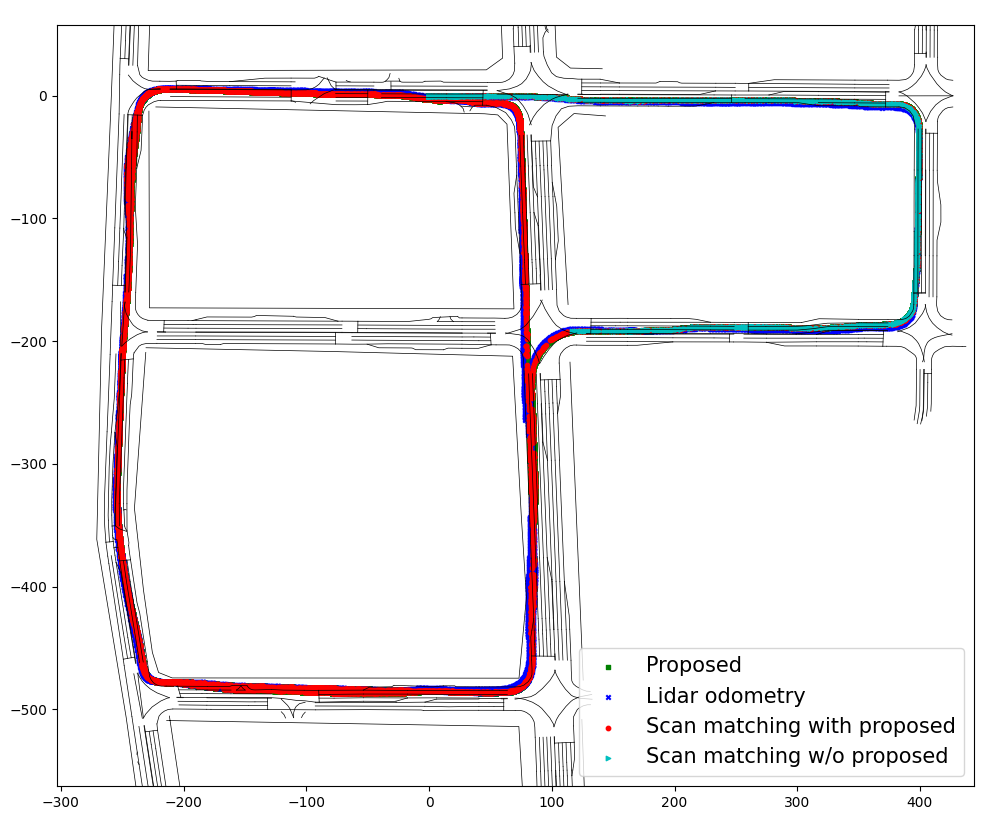}
        \label{fig:localize_1}
    }
    \subfigure[]{
        \includegraphics[width=0.9\columnwidth]{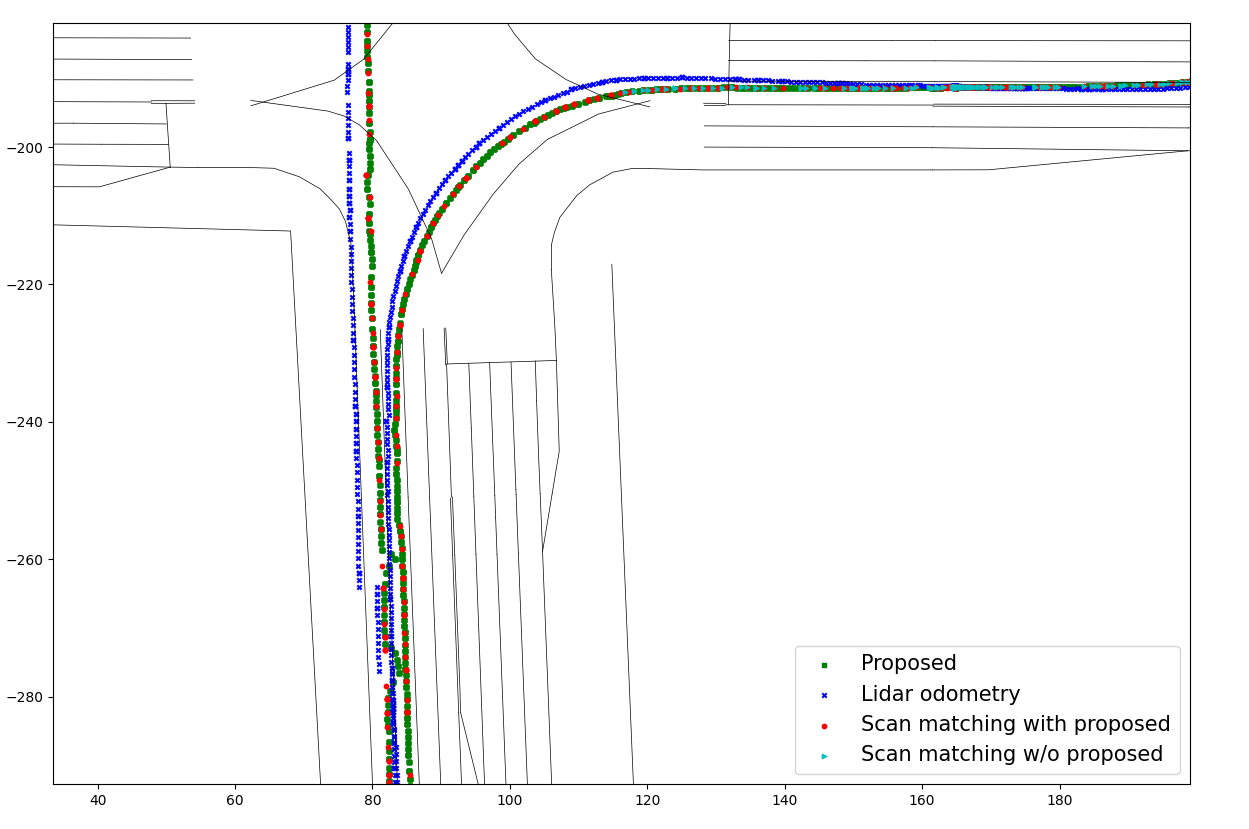}
        \label{fig:localize_2}
    }
    \caption{
    (a) The result of localization methods. Our proposed method can estimate the pose in an urban area utilizing LiDAR odometry and scan-matching resiliently. 
    (b) LiDAR odometry shows a drift on the straightaway. In addition, the single scan-matching method without our proposed method diverges after a left-turn at the intersection.
    \label{fig:localization}
    }
\end{figure}
The results of the resilient localization algorithm are depicted in Fig. \ref{fig:localization}. 
Our proposed methods can estimate the pose in the urban area utilizing LiDAR odometry and scan-matching resiliently. 
Our proposed localization system enhances accuracy and resilience using sliding-window-based scan-matching and LiDAR odometry-based prediction.
\begin{table}[hbt!]
\caption[]{Proposed localization results. The proposed method has the lowest average error while still maintaining a high frequency. A combination of scan matching and LiDAR odometry was utilized for this purpose. Additionally, the scan matching algorithm was improved using the proposed method. In cases where only scan matching is used, scan matching failed when the vehicle was driven dynamically.}
\label{tab:localize_compare}
\centering
\begin{tabular}{c|cccc}
\hline
Method                                                                    &
\textbf{Proposed}                                                                  & 
\begin{tabular}[c]{@{}c@{}}LiDAR \\ odometry \end{tabular}                &
\begin{tabular}[c]{@{}c@{}}Scan match \\ (w. proposed) \end{tabular}      &
\begin{tabular}[c]{@{}c@{}}Scan match \\ (w/o. proposed) \end{tabular} 
\\ \hline
\begin{tabular}[c]{@{}c@{}}Average \\ error(m) \end{tabular}              &
\textbf{0.05678}                                                                    & 
1.0162                                                                    & 
0.0314                                                         & 
Fail     
\\ \hline
\begin{tabular}[c]{@{}c@{}}Number \\ of data \end{tabular}              &
\textbf{82,137}                                                                    & 
9,157                                                                    & 
6,399                                                         & 
1,943     
\\ \hline
\end{tabular}
\end{table}
The accuracy results are listed in Table \ref{tab:localize_compare}.
As illustrated in Fig. \ref{fig:localize_2}, the conventional scan-matching method without our proposed methods diverges after a left-turn at the intersection.
However, utilizing a robust localization algorithm, we can retain the performance of the scan-matching algorithm covering the entire map, whereas the conventional matching frequency was low. 
Moreover, LiDAR odometry is also able to cover the entire map, but it has a higher average error in the large-scale map.
Therefore, we verified that the proposed algorithm was able to estimate the vehicle pose  with high-frequency covering the speed up to 60 $kph$ without using a GPS.
In addition, we validated our proposed method in an area of over 400,000 $m^2$ and a course length of approximately 5 $km$.

\subsection{Motion planning}
We have proposed road-graph searching-based planning and motion primitives-based planning for macro and micro motion planning algorithms, respectively.
In addition, as we implemented the TSPS algorithm to select an optimal path according to driving situations, 
we designed our motion planning system to find a collision-free path utilizing a road-graph and motion primitives.
\begin{figure}[!t]
    \centering
    \subfigure{
        \includegraphics[width=1.0\columnwidth]{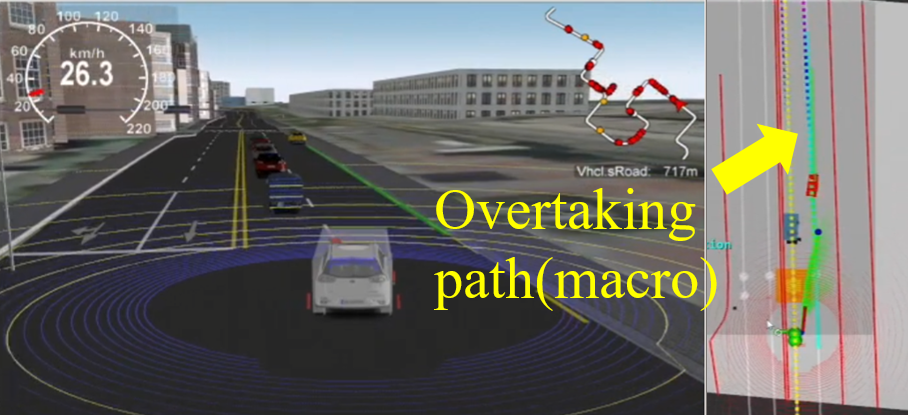}
    }
    \subfigure{
        \includegraphics[width=1.0\columnwidth]{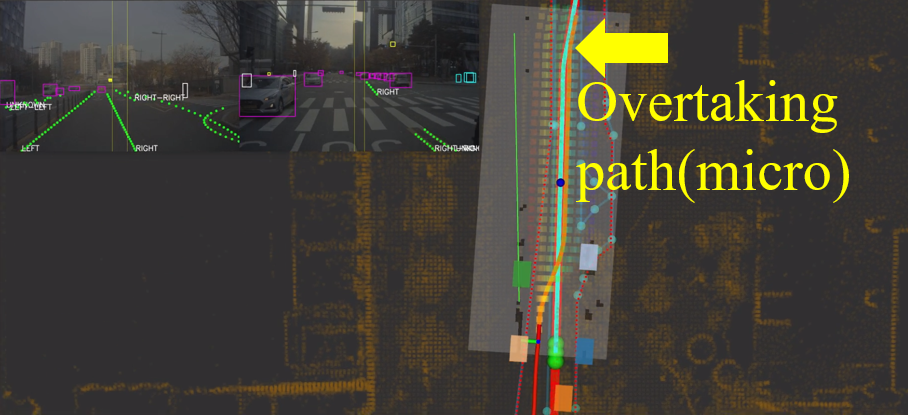}
    }
    \caption{
    The proposed overtaking trajectory was visualized. A collision-free path can be generated by combining macro and micro overtaking paths. 
    (a) Using the simulation environment, macro overtaking paths are visualized. 
    (b) The macro overtaking path can be adjusted based on motion primitives if the micro overtaking path is enabled. We conducted this evaluation in the real world. 
    \label{fig:motion_planning_result}
    }
\end{figure}
As a result, we evaluated our motion planning algorithm in the simulation environment as well as in real-world experiments as illustrated in Fig. \ref{fig:motion_planning_result}.
Furthermore, in the competition event, we validated our algorithm that enabled our vehicle to overtake the congested traffic situation as shown in Fig. \ref{fig:real_world_overtaking}.

\begin{figure}[ht]
    \centering
    \includegraphics[width=1\columnwidth]{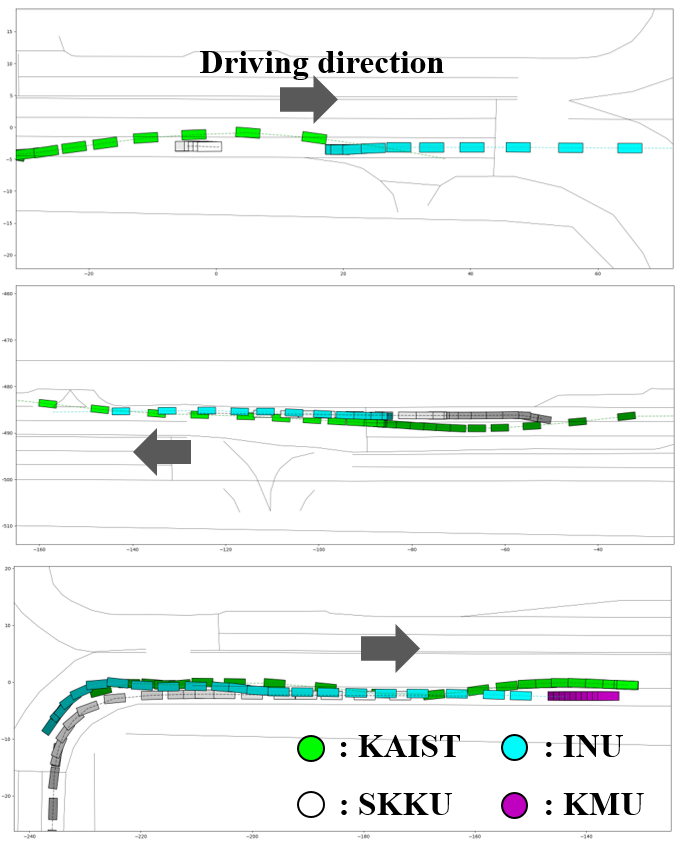}
    \caption{In the real world, there are overtaking scenarios during the final competition. 
    To visualize the history trajectory, we used consecutive colored boxes. The vehicle computed overtaking paths in a congested environment.}
    \label{fig:real_world_overtaking}
\end{figure}

\subsection{Behavior planning}
\begin{figure}[ht]
\centering
\includegraphics[width=0.98\columnwidth]{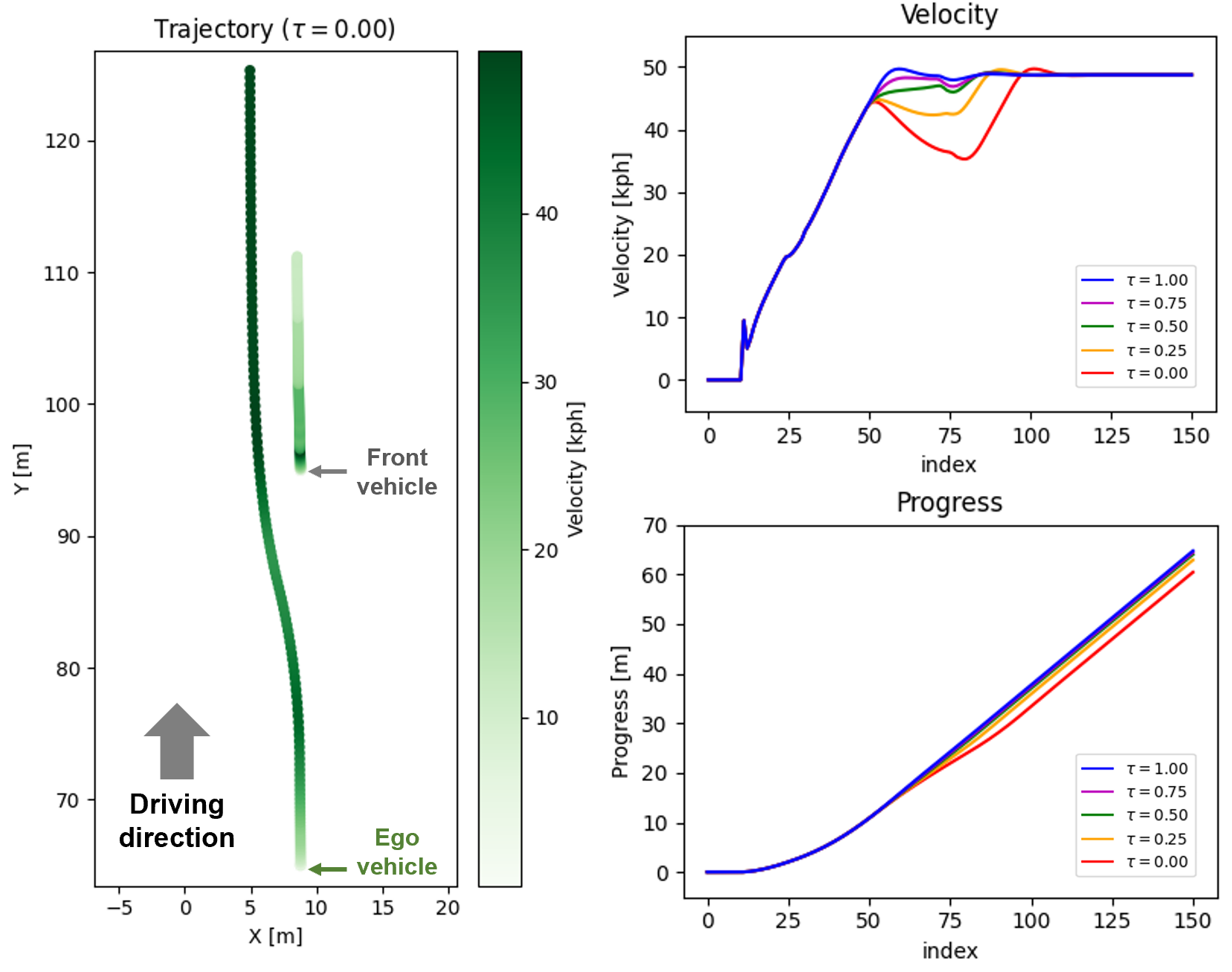}
\caption{
The driven trajectory with the proposed geometry-aware velocity planning algorithm in overtaking scenario (left).
The results of the velocity (right-top) and progress (right-bottom) show that the aggressiveness factor $\tau$ can change the longitudinal maneuver of the ego vehicle in the velocity and progress perspectives.
}
\label{fig:GVP_result}
\end{figure}

Before real-world deployment, we studied the quantitative evaluation of the TSPS and GVP algorithms in the simulated traffic environment with the various aggressiveness factors $\tau$. 
Because the factor can change the resultant behavior of the ego vehicle, we investigated the lateral and longitudinal maneuvers concerning five different aggressiveness factors. 
Figure \ref{fig:GVP_result} (left) shows a driven trajectory of the ego and front vehicles during an overtaking scenario. While the TSPS generated collision-free motion plans, the GVP computes proper velocity plans considering the dynamic geometric relationship during the lateral overtaking maneuver. 
We further evaluated our algorithms with longitudinal metrics, such as velocity and progress.
Because a lower $\tau$ makes the ego vehicle conservative, the most considerable velocity and terminal progress drop was at $\tau=0$. By contrast, when $\tau=1$, the ego vehicle performed overtaking with almost no velocity reduction and showed the highest terminal progress.
Considering the simulated environment allows perfect observation compared to the real world, we choose $\tau=0.75$ which shows a minor velocity drop and progress decrement in real-world deployment. 

\begin{table}[hbt!]
\caption[The results of the spatial similarity comparison based on Kullback-Leibler divergence (KLD) and mean error.]{The results of the spatial similarity comparison based on Kullback-Leibler divergence (KLD) and mean error.}
\label{tab:full_race_traffic}
\centering
\begin{tabular}{c|ccccc}
\hline
Team & KAIST & CBNU & INU & INHA & SKKU \\ 
\hline
\hline
KAIST & 
    \begin{tabular}[c]{@{}c@{}} - (KLD)\\ - (Mean) \end{tabular} & 
    \begin{tabular}[c]{@{}c@{}}13.343\\ 12.461 \end{tabular} & 
    \begin{tabular}[c]{@{}c@{}}14.286\\ 2.21 \end{tabular} &
    \begin{tabular}[c]{@{}c@{}}13.452\\ 13.037 \end{tabular} & 
    \begin{tabular}[c]{@{}c@{}}14.155\\ 1.58 \end{tabular} \\
    \hline
CBNU  & 
    \begin{tabular}[c]{@{}c@{}} -\\ - \end{tabular} & 
    \begin{tabular}[c]{@{}c@{}} -\\ - \end{tabular} & 
    \begin{tabular}[c]{@{}c@{}}\textbf{0.943}\\ \textbf{1.764} \end{tabular} &
    \begin{tabular}[c]{@{}c@{}}\textbf{0.109}\\ 6.209 \end{tabular} & 
    \begin{tabular}[c]{@{}c@{}}\textbf{0.812}\\ 1.569 \end{tabular} \\
    \hline
INU   & 
    \begin{tabular}[c]{@{}c@{}} -\\ - \end{tabular} & 
    \begin{tabular}[c]{@{}c@{}} -\\ - \end{tabular} & 
    \begin{tabular}[c]{@{}c@{}} -\\ - \end{tabular} &
    \begin{tabular}[c]{@{}c@{}}\textbf{0.833}\\ 10.016 \end{tabular} &
    \begin{tabular}[c]{@{}c@{}}\textbf{0.131}\\ 0.792 \end{tabular} \\
    \hline
INHA  & 
    \begin{tabular}[c]{@{}c@{}} -\\ - \end{tabular} & 
    \begin{tabular}[c]{@{}c@{}} -\\ - \end{tabular} & 
    \begin{tabular}[c]{@{}c@{}} -\\ - \end{tabular} &
    \begin{tabular}[c]{@{}c@{}} -\\ - \end{tabular} & 
    \begin{tabular}[c]{@{}c@{}}\textbf{0.702}\\ 1.158 \end{tabular} \\
    \hline

\end{tabular}
\end{table}

\begin{table}[hbt!]
\caption[Results of curve section-specific comparisons based on RMSE and KLD.]{Results of curve section-specific comparisons based on RMSE and KLD.}
\label{tab:curve_1}
\centering
\begin{tabular}{c|c|ccccc}
\hline
\multicolumn{2}{c}{Team} & KAIST & CBNU & INU & INHA & SKKU \\ 
\hline
\hline
\multirow{8}{*}{\begin{turn}{90}curve A\end{turn}}
& KAIST & 
    \begin{tabular}[c]{@{}c@{}} - (KLD)\\ - (Mean) \end{tabular} & 
    \begin{tabular}[c]{@{}c@{}}  13.211 \\  1.977 \end{tabular} & 
    \begin{tabular}[c]{@{}c@{}}  14.396 \\  1.896 \end{tabular} & 
    \begin{tabular}[c]{@{}c@{}}  19.807 \\  2.280 \end{tabular} & 
    \begin{tabular}[c]{@{}c@{}}  11.741 \\  2.126 \end{tabular} \\
    \cline{2-7}
& CBNU & 
    \begin{tabular}[c]{@{}c@{}} -\\ - \end{tabular} & 
    \begin{tabular}[c]{@{}c@{}} -\\ - \end{tabular} & 
    \begin{tabular}[c]{@{}c@{}}  0.452 \\ 0.497 \end{tabular} & 
    \begin{tabular}[c]{@{}c@{}}  3.946 \\  0.872 \end{tabular} & 
    \begin{tabular}[c]{@{}c@{}}  0.924 \\  0.350 \end{tabular} \\
    \cline{2-7}
& INU & 
    \begin{tabular}[c]{@{}c@{}} -\\ - \end{tabular} & 
    \begin{tabular}[c]{@{}c@{}} -\\ - \end{tabular} & 
    \begin{tabular}[c]{@{}c@{}} -\\ - \end{tabular} &
    \begin{tabular}[c]{@{}c@{}}  3.230 \\  0.849 \end{tabular} & 
    \begin{tabular}[c]{@{}c@{}}  1.586 \\  0.737 \end{tabular} \\
    \cline{2-7}
& INHA & 
    \begin{tabular}[c]{@{}c@{}} -\\ - \end{tabular} & 
    \begin{tabular}[c]{@{}c@{}} -\\ - \end{tabular} & 
    \begin{tabular}[c]{@{}c@{}} -\\ - \end{tabular} &
    \begin{tabular}[c]{@{}c@{}} -\\ - \end{tabular} & 
    \begin{tabular}[c]{@{}c@{}} 5.306 \\ 0.844 \end{tabular} \\
    \hline
    \hline
\multirow{8}{*}{\begin{turn}{90}curve B\end{turn}}
& KAIST & 
    \begin{tabular}[c]{@{}c@{}} - (KLD)\\ - (Mean) \end{tabular} & 
    \begin{tabular}[c]{@{}c@{}}  13.769 \\ 1.699 \end{tabular} & 
    \begin{tabular}[c]{@{}c@{}}  13.611 \\ 1.838 \end{tabular} & 
    \begin{tabular}[c]{@{}c@{}}  13.660 \\ 1.629 \end{tabular} & 
    \begin{tabular}[c]{@{}c@{}}  17.508 \\ 2.535 \end{tabular}\\
    \cline{2-7}
& CBNU  & 
    \begin{tabular}[c]{@{}c@{}} -\\ - \end{tabular} & 
    \begin{tabular}[c]{@{}c@{}} -\\ - \end{tabular} & 
    \begin{tabular}[c]{@{}c@{}} 0.011 \\ 0.260 \end{tabular} & 
    \begin{tabular}[c]{@{}c@{}} 0.398 \\ 0.860 \end{tabular} & 
    \begin{tabular}[c]{@{}c@{}} 2.407 \\ 0.352 \end{tabular} \\
    \cline{2-7}
& INU   & 
    \begin{tabular}[c]{@{}c@{}} -\\ - \end{tabular} & 
    \begin{tabular}[c]{@{}c@{}} -\\ - \end{tabular} & 
    \begin{tabular}[c]{@{}c@{}} -\\ - \end{tabular} &
    \begin{tabular}[c]{@{}c@{}} 0.375 \\ 0.734 \end{tabular} & 
    \begin{tabular}[c]{@{}c@{}} 2.863 \\ 0.539 \end{tabular} \\
    \cline{2-7}
& INHA  & 
    \begin{tabular}[c]{@{}c@{}} -\\ - \end{tabular} & 
    \begin{tabular}[c]{@{}c@{}} -\\ - \end{tabular} & 
    \begin{tabular}[c]{@{}c@{}} -\\ - \end{tabular} &
    \begin{tabular}[c]{@{}c@{}} -\\ - \end{tabular} & 
    \begin{tabular}[c]{@{}c@{}} 3.604 \\ 1.180 \end{tabular}  \\
    \hline
    \hline

\multirow{8}{*}{\begin{turn}{90}curve D(Lap 1)\end{turn}}
& KAIST & 
    \begin{tabular}[c]{@{}c@{}} - (KLD)\\ - (Mean) \end{tabular} & 
    \begin{tabular}[c]{@{}c@{}}  15.506 \\ 1.433 \end{tabular} & 
    \begin{tabular}[c]{@{}c@{}}  26.581 \\ 1.676 \end{tabular} & 
    \begin{tabular}[c]{@{}c@{}}  22.547 \\ 1.169 \end{tabular} & 
    \begin{tabular}[c]{@{}c@{}}  27.501 \\ 1.563 \end{tabular} \\
    \cline{2-7}
& CBNU  & 
    \begin{tabular}[c]{@{}c@{}} -\\ - \end{tabular} & 
    \begin{tabular}[c]{@{}c@{}} -\\ - \end{tabular} & 
    \begin{tabular}[c]{@{}c@{}}  7.073 \\ 0.712 \end{tabular} & 
    \begin{tabular}[c]{@{}c@{}}  3.750 \\ 0.559 \end{tabular} & 
    \begin{tabular}[c]{@{}c@{}}  8.588 \\ 0.582 \end{tabular} \\
    \cline{2-7}
& INU   & 
    \begin{tabular}[c]{@{}c@{}} -\\ - \end{tabular} & 
    \begin{tabular}[c]{@{}c@{}} -\\ - \end{tabular} & 
    \begin{tabular}[c]{@{}c@{}} -\\ - \end{tabular} &
    \begin{tabular}[c]{@{}c@{}}  2.827 \\ 0.470 \end{tabular} & 
    \begin{tabular}[c]{@{}c@{}}  0.161 \\ 0.615 \end{tabular} \\
    \cline{2-7}
& INHA  & 
    \begin{tabular}[c]{@{}c@{}} -\\ - \end{tabular} & 
    \begin{tabular}[c]{@{}c@{}} -\\ - \end{tabular} & 
    \begin{tabular}[c]{@{}c@{}} -\\ - \end{tabular} &
    \begin{tabular}[c]{@{}c@{}} -\\ - \end{tabular} & 
    \begin{tabular}[c]{@{}c@{}}  3.399 \\ 0.799 \end{tabular} \\
    \hline
    \hline

\multirow{8}{*}{\begin{turn}{90}curve D (Lap 2)\end{turn}}
& KAIST & 
    \begin{tabular}[c]{@{}c@{}} - (KLD)\\ - (Mean) \end{tabular} & 
    \begin{tabular}[c]{@{}c@{}}  11.102 \\ 1.870 \end{tabular} & 
    \begin{tabular}[c]{@{}c@{}}  12.517 \\ 1.648 \end{tabular} & 
    \begin{tabular}[c]{@{}c@{}}  19.534 \\ 2.854 \end{tabular} & 
    \begin{tabular}[c]{@{}c@{}}  15.618 \\ 1.862 \end{tabular} \\
    \cline{2-7}
& CBNU  & 
    \begin{tabular}[c]{@{}c@{}} -\\ - \end{tabular} & 
    \begin{tabular}[c]{@{}c@{}} -\\ - \end{tabular} & 
    \begin{tabular}[c]{@{}c@{}}  0.601 \\ 0.474 \end{tabular} & 
    \begin{tabular}[c]{@{}c@{}}  8.581 \\ 1.721 \end{tabular} & 
    \begin{tabular}[c]{@{}c@{}}  3.158 \\ 0.354 \end{tabular} \\
    \cline{2-7}
& INU   & 
    \begin{tabular}[c]{@{}c@{}} -\\ - \end{tabular} & 
    \begin{tabular}[c]{@{}c@{}} -\\ - \end{tabular} & 
    \begin{tabular}[c]{@{}c@{}} -\\ - \end{tabular} &
    \begin{tabular}[c]{@{}c@{}}  7.983 \\ 1.606 \end{tabular} & 
    \begin{tabular}[c]{@{}c@{}}  2.436 \\ 0.728 \end{tabular} \\
    \cline{2-7}
& INHA  & 
    \begin{tabular}[c]{@{}c@{}} -\\ - \end{tabular} & 
    \begin{tabular}[c]{@{}c@{}} -\\ - \end{tabular} & 
    \begin{tabular}[c]{@{}c@{}} -\\ - \end{tabular} &
    \begin{tabular}[c]{@{}c@{}} -\\ - \end{tabular} & 
    \begin{tabular}[c]{@{}c@{}} 5.539 \\ 1.696 \end{tabular} \\
    \hline
    \hline

\multirow{8}{*}{\begin{turn}{90}curve E (Lap 1)\end{turn}}
& KAIST & 
    \begin{tabular}[c]{@{}c@{}} - (KLD)\\ - (Mean) \end{tabular} & 
    \begin{tabular}[c]{@{}c@{}}  15.278 \\ 2.194 \end{tabular} & 
    \begin{tabular}[c]{@{}c@{}}  14.406 \\ 2.279 \end{tabular} & 
    \begin{tabular}[c]{@{}c@{}}  22.287 \\ 1.682 \end{tabular} & 
    \begin{tabular}[c]{@{}c@{}}  22.113 \\ 2.534 \end{tabular} \\
    \cline{2-7}
& CBNU  & 
    \begin{tabular}[c]{@{}c@{}} -\\ - \end{tabular} & 
    \begin{tabular}[c]{@{}c@{}} -\\ - \end{tabular} & 
    \begin{tabular}[c]{@{}c@{}}  0.405 \\ 0.239 \end{tabular} & 
    \begin{tabular}[c]{@{}c@{}}  6.006 \\ 0.483 \end{tabular} & 
    \begin{tabular}[c]{@{}c@{}}  4.993 \\ 0.608 \end{tabular} \\
    \cline{2-7}
& INU   & 
    \begin{tabular}[c]{@{}c@{}} -\\ - \end{tabular} & 
    \begin{tabular}[c]{@{}c@{}} -\\ - \end{tabular} & 
    \begin{tabular}[c]{@{}c@{}} -\\ - \end{tabular} &
    \begin{tabular}[c]{@{}c@{}}  6.472 \\ 0.616 \end{tabular} & 
    \begin{tabular}[c]{@{}c@{}}  5.455 \\ 0.807 \end{tabular} \\
    \cline{2-7}
& INHA  & 
    \begin{tabular}[c]{@{}c@{}} -\\ - \end{tabular} & 
    \begin{tabular}[c]{@{}c@{}} -\\ - \end{tabular} & 
    \begin{tabular}[c]{@{}c@{}} -\\ - \end{tabular} &
    \begin{tabular}[c]{@{}c@{}} -\\ - \end{tabular} & 
    \begin{tabular}[c]{@{}c@{}} 0.846 \\ 1.046 \end{tabular} \\
    \hline
\end{tabular}
\end{table}

\begin{table}[hbt!]
\caption[Results of curve section-specific comparisons based on RMSE and KLD.]{Results of curve section-specific comparisons based on RMSE and KLD.}
\label{tab:curve_2}
\centering
\begin{tabular}{c|c|ccccc}
\hline
\multicolumn{2}{c}{Team} & KAIST & CBNU & INU & INHA & SKKU \\ 
\hline
\hline
\multirow{8}{*}{\begin{turn}{90}curve E (Lap 2)\end{turn}}
& KAIST & 
    \begin{tabular}[c]{@{}c@{}} - (KLD)\\ - (Mean) \end{tabular} & 
    \begin{tabular}[c]{@{}c@{}}  19.076 \\ 2.046 \end{tabular} & 
    \begin{tabular}[c]{@{}c@{}}  15.987 \\ 2.194 \end{tabular} & 
    \begin{tabular}[c]{@{}c@{}}  14.185 \\ 1.685 \end{tabular} & 
    \begin{tabular}[c]{@{}c@{}}  22.591 \\ 2.438 \end{tabular} \\
    \cline{2-7}
& CBNU  & 
    \begin{tabular}[c]{@{}c@{}} -\\ - \end{tabular} & 
    \begin{tabular}[c]{@{}c@{}} -\\ - \end{tabular} & 
    \begin{tabular}[c]{@{}c@{}}  2.715 \\ 0.246 \end{tabular} & 
    \begin{tabular}[c]{@{}c@{}}  4.703 \\ 0.490 \end{tabular} & 
    \begin{tabular}[c]{@{}c@{}}  1.440 \\ 0.698 \end{tabular} \\
    \cline{2-7}
& INU   & 
    \begin{tabular}[c]{@{}c@{}} -\\ - \end{tabular} & 
    \begin{tabular}[c]{@{}c@{}} -\\ - \end{tabular} & 
    \begin{tabular}[c]{@{}c@{}} -\\ - \end{tabular} &
    \begin{tabular}[c]{@{}c@{}}  2.004 \\ 0.616 \end{tabular} & 
    \begin{tabular}[c]{@{}c@{}}  4.278 \\ 0.906 \end{tabular} \\
    \cline{2-7}
& INHA  & 
    \begin{tabular}[c]{@{}c@{}} -\\ - \end{tabular} & 
    \begin{tabular}[c]{@{}c@{}} -\\ - \end{tabular} & 
    \begin{tabular}[c]{@{}c@{}} -\\ - \end{tabular} &
    \begin{tabular}[c]{@{}c@{}} -\\ - \end{tabular} & 
    \begin{tabular}[c]{@{}c@{}}  6.537 \\ 0.935 \end{tabular} \\
    \hline
    \hline
\multirow{8}{*}{\begin{turn}{90}curve F (Lap 1)\end{turn}}
& KAIST & 
    \begin{tabular}[c]{@{}c@{}} - (KLD)\\ - (Mean) \end{tabular} & 
    \begin{tabular}[c]{@{}c@{}}  38.144 \\ 3.172 \end{tabular} & 
    \begin{tabular}[c]{@{}c@{}}  12.105 \\ 2.010 \end{tabular} & 
    \begin{tabular}[c]{@{}c@{}}  17.878 \\ 1.914 \end{tabular} & 
    \begin{tabular}[c]{@{}c@{}}  10.080 \\ 1.982 \end{tabular} \\
    \cline{2-7}
& CBNU  & 
    \begin{tabular}[c]{@{}c@{}} -\\ - \end{tabular} & 
    \begin{tabular}[c]{@{}c@{}} -\\ - \end{tabular} & 
    \begin{tabular}[c]{@{}c@{}} 22.991 \\ 1.681 \end{tabular} & 
    \begin{tabular}[c]{@{}c@{}} 16.946 \\ 1.035 \end{tabular} & 
    \begin{tabular}[c]{@{}c@{}} 25.761 \\ 1.852 \end{tabular} \\
    \cline{2-7}
& INU   & 
    \begin{tabular}[c]{@{}c@{}} -\\ - \end{tabular} & 
    \begin{tabular}[c]{@{}c@{}} -\\ - \end{tabular} & 
    \begin{tabular}[c]{@{}c@{}} -\\ - \end{tabular} &
    \begin{tabular}[c]{@{}c@{}}  5.037 \\ 0.593 \end{tabular} & 
    \begin{tabular}[c]{@{}c@{}}  1.783 \\ 0.367 \end{tabular} \\
    \cline{2-7}
& INHA  & 
    \begin{tabular}[c]{@{}c@{}} -\\ - \end{tabular} & 
    \begin{tabular}[c]{@{}c@{}} -\\ - \end{tabular} & 
    \begin{tabular}[c]{@{}c@{}} -\\ - \end{tabular} &
    \begin{tabular}[c]{@{}c@{}} -\\ - \end{tabular} & 
    \begin{tabular}[c]{@{}c@{}}  6.987 \\ 0.857 \end{tabular} \\
    \hline
    \hline

\multirow{8}{*}{\begin{turn}{90}curve F (Lap 2)\end{turn}}
& KAIST & 
    \begin{tabular}[c]{@{}c@{}} - (KLD)\\ - (Mean) \end{tabular} & 
    \begin{tabular}[c]{@{}c@{}}  28.006 \\ 0.781 \end{tabular} & 
    \begin{tabular}[c]{@{}c@{}}  0.863 \\ 1.156 \end{tabular} & 
    \begin{tabular}[c]{@{}c@{}}  13.113 \\ 1.513 \end{tabular} & 
    \begin{tabular}[c]{@{}c@{}}  13.118 \\ 1.160 \end{tabular} \\
    \cline{2-7}
& CBNU  & 
    \begin{tabular}[c]{@{}c@{}} -\\ - \end{tabular} & 
    \begin{tabular}[c]{@{}c@{}} -\\ - \end{tabular} & 
    \begin{tabular}[c]{@{}c@{}} 27.039 \\ 2.050 \end{tabular} & 
    \begin{tabular}[c]{@{}c@{}} 12.791 \\ 0.777 \end{tabular} & 
    \begin{tabular}[c]{@{}c@{}} 13.329 \\ 0.235 \end{tabular} \\
    \cline{2-7}
& INU   & 
    \begin{tabular}[c]{@{}c@{}} -\\ - \end{tabular} & 
    \begin{tabular}[c]{@{}c@{}} -\\ - \end{tabular} & 
    \begin{tabular}[c]{@{}c@{}} -\\ - \end{tabular} &
    \begin{tabular}[c]{@{}c@{}}  12.078 \\ 1.822 \end{tabular} & 
    \begin{tabular}[c]{@{}c@{}}  11.580 \\ 1.686 \end{tabular} \\
    \cline{2-7}
& INHA  & 
    \begin{tabular}[c]{@{}c@{}} -\\ - \end{tabular} & 
    \begin{tabular}[c]{@{}c@{}} -\\ - \end{tabular} & 
    \begin{tabular}[c]{@{}c@{}} -\\ - \end{tabular} &
    \begin{tabular}[c]{@{}c@{}} -\\ - \end{tabular} & 
    \begin{tabular}[c]{@{}c@{}} 0.420 \\ 0.712 \end{tabular} \\
    \hline
    \hline

\multirow{8}{*}{\begin{turn}{90}curve G (Lap 1)\end{turn}}
& KAIST & 
    \begin{tabular}[c]{@{}c@{}} - (KLD)\\ - (Mean) \end{tabular} & 
    \begin{tabular}[c]{@{}c@{}}  2.760 \\ 0.919 \end{tabular} & 
    \begin{tabular}[c]{@{}c@{}}  4.893 \\ 0.770 \end{tabular} & 
    \begin{tabular}[c]{@{}c@{}}  9.083 \\ 1.037 \end{tabular} & 
    \begin{tabular}[c]{@{}c@{}}  1.195 \\ 1.034 \end{tabular} \\
    \cline{2-7}
& CBNU  & 
    \begin{tabular}[c]{@{}c@{}} -\\ - \end{tabular} & 
    \begin{tabular}[c]{@{}c@{}} -\\ - \end{tabular} & 
    \begin{tabular}[c]{@{}c@{}}  8.889 \\ 0.279 \end{tabular} & 
    \begin{tabular}[c]{@{}c@{}}  11.788 \\ 0.728 \end{tabular} & 
    \begin{tabular}[c]{@{}c@{}}  3.924 \\ 0.419 \end{tabular} \\
    \cline{2-7}
& INU   & 
    \begin{tabular}[c]{@{}c@{}} -\\ - \end{tabular} & 
    \begin{tabular}[c]{@{}c@{}} -\\ - \end{tabular} & 
    \begin{tabular}[c]{@{}c@{}} -\\ - \end{tabular} &
    \begin{tabular}[c]{@{}c@{}}  3.093 \\ 0.919 \end{tabular} & 
    \begin{tabular}[c]{@{}c@{}}  4.785 \\ 0.249 \end{tabular} \\
    \cline{2-7}
& INHA  & 
    \begin{tabular}[c]{@{}c@{}} -\\ - \end{tabular} & 
    \begin{tabular}[c]{@{}c@{}} -\\ - \end{tabular} & 
    \begin{tabular}[c]{@{}c@{}} -\\ - \end{tabular} &
    \begin{tabular}[c]{@{}c@{}} -\\ - \end{tabular} & 
    \begin{tabular}[c]{@{}c@{}} 7.905 \\ 0.871 \end{tabular} \\
    \hline
    \hline

\multirow{8}{*}{\begin{turn}{90}curve G (Lap 2)\end{turn}}
& KAIST & 
    \begin{tabular}[c]{@{}c@{}} - (KLD)\\ - (Mean) \end{tabular} & 
    \begin{tabular}[c]{@{}c@{}}  6.751 \\ 0.948 \end{tabular} & 
    \begin{tabular}[c]{@{}c@{}}  12.259 \\ 4.320 \end{tabular} & 
    \begin{tabular}[c]{@{}c@{}}  9.260 \\ 1.858 \end{tabular} & 
    \begin{tabular}[c]{@{}c@{}}  7.881 \\ 0.949 \end{tabular} \\
    \cline{2-7}
& CBNU  & 
    \begin{tabular}[c]{@{}c@{}} -\\ - \end{tabular} & 
    \begin{tabular}[c]{@{}c@{}} -\\ - \end{tabular} & 
    \begin{tabular}[c]{@{}c@{}}  6.049 \\ 3.247 \end{tabular} & 
    \begin{tabular}[c]{@{}c@{}}  2.916 \\ 1.767 \end{tabular} & 
    \begin{tabular}[c]{@{}c@{}}  0.794 \\ 0.469 \end{tabular} \\
    \cline{2-7}
& INU   & 
    \begin{tabular}[c]{@{}c@{}} -\\ - \end{tabular} & 
    \begin{tabular}[c]{@{}c@{}} -\\ - \end{tabular} & 
    \begin{tabular}[c]{@{}c@{}} -\\ - \end{tabular} &
    \begin{tabular}[c]{@{}c@{}} 3.161 \\ 2.826 \end{tabular} & 
    \begin{tabular}[c]{@{}c@{}} 5.566 \\ 3.723 \end{tabular} \\
    \cline{2-7}
& INHA  & 
    \begin{tabular}[c]{@{}c@{}} -\\ - \end{tabular} & 
    \begin{tabular}[c]{@{}c@{}} -\\ - \end{tabular} & 
    \begin{tabular}[c]{@{}c@{}} -\\ - \end{tabular} &
    \begin{tabular}[c]{@{}c@{}} -\\ - \end{tabular} & 
    \begin{tabular}[c]{@{}c@{}} 2.187 \\ 1.585 \end{tabular} \\
    \hline
    \hline

\end{tabular}
\end{table}

\begin{table}[hbt!]
\caption[Results of intersection-specific comparisons based on RMSE and KLD.]{Results of intersection-specific comparisons based on RMSE and KLD.}
\label{tab:intersection}
\centering
\begin{tabular}{c|c|ccccc}
\hline
\multicolumn{2}{c}{Team} & KAIST & CBNU & INU & INHA & SKKU \\ 
\hline
\hline
\multirow{8}{*}{\begin{turn}{90}intersection C\end{turn}}
& KAIST & 
    \begin{tabular}[c]{@{}c@{}} - (KLD)\\ - (Mean) \end{tabular} & 
    \begin{tabular}[c]{@{}c@{}}  7.860 \\ 2.558 \end{tabular} & 
    \begin{tabular}[c]{@{}c@{}}  9.057 \\ 1.651 \end{tabular} & 
    \begin{tabular}[c]{@{}c@{}}  4.588 \\ 1.066 \end{tabular} & 
    \begin{tabular}[c]{@{}c@{}}  8.014 \\ 1.564 \end{tabular} \\
    \cline{2-7}
& CBNU  & 
    \begin{tabular}[c]{@{}c@{}} -\\ - \end{tabular} & 
    \begin{tabular}[c]{@{}c@{}} -\\ - \end{tabular} & 
    \begin{tabular}[c]{@{}c@{}}  1.000 \\ 4.360 \end{tabular} & 
    \begin{tabular}[c]{@{}c@{}}  3.345 \\ 3.109 \end{tabular} & 
    \begin{tabular}[c]{@{}c@{}}  0.207 \\ 2.920 \end{tabular} \\
    \cline{2-7}
& INU   & 
    \begin{tabular}[c]{@{}c@{}} -\\ - \end{tabular} & 
    \begin{tabular}[c]{@{}c@{}} -\\ - \end{tabular} & 
    \begin{tabular}[c]{@{}c@{}} -\\ - \end{tabular} &
    \begin{tabular}[c]{@{}c@{}} 4.534 \\ 1.938 \end{tabular} & 
    \begin{tabular}[c]{@{}c@{}} 0.752 \\ 0.900 \end{tabular} \\
    \cline{2-7}
& INHA  & 
    \begin{tabular}[c]{@{}c@{}} -\\ - \end{tabular} & 
    \begin{tabular}[c]{@{}c@{}} -\\ - \end{tabular} & 
    \begin{tabular}[c]{@{}c@{}} -\\ - \end{tabular} &
    \begin{tabular}[c]{@{}c@{}} -\\ - \end{tabular} & 
    \begin{tabular}[c]{@{}c@{}}  3.624 \\ 1.803 \end{tabular} \\
    \hline

\end{tabular}
\end{table}

\begin{table}[hbt!]
\caption[Results of straight section-specific comparisons based on RMSE and KLD.]{Results of straight section-specific comparisons based on RMSE and KLD.}
\label{tab:straight}
\centering
\begin{tabular}{c|c|ccccc}
\hline
\multicolumn{2}{c}{Team} & KAIST & CBNU & INU & INHA & SKKU \\ 
\hline
\hline
\multirow{8}{*}{\begin{turn}{90}straight H\end{turn}}
& KAIST & 
    \begin{tabular}[c]{@{}c@{}} - (KLD)\\ - (Mean) \end{tabular} & 
    \begin{tabular}[c]{@{}c@{}}  25.679 \\ 2.281 \end{tabular} & 
    \begin{tabular}[c]{@{}c@{}}  14.105 \\ 0.883 \end{tabular} & 
    \begin{tabular}[c]{@{}c@{}}  21.677 \\ 0.979 \end{tabular} & 
    \begin{tabular}[c]{@{}c@{}}  15.267 \\ 1.055 \end{tabular} \\
    \cline{2-7}
& CBNU  & 
    \begin{tabular}[c]{@{}c@{}} -\\ - \end{tabular} & 
    \begin{tabular}[c]{@{}c@{}} -\\ - \end{tabular} & 
    \begin{tabular}[c]{@{}c@{}} 10.948 \\ 0.492 \end{tabular} & 
    \begin{tabular}[c]{@{}c@{}} 4.272 \\ 0.362 \end{tabular} & 
    \begin{tabular}[c]{@{}c@{}} 9.795 \\ 0.457 \end{tabular}  \\
    \cline{2-7}
& INU   & 
    \begin{tabular}[c]{@{}c@{}} -\\ - \end{tabular} & 
    \begin{tabular}[c]{@{}c@{}} -\\ - \end{tabular} & 
    \begin{tabular}[c]{@{}c@{}} -\\ - \end{tabular} &
    \begin{tabular}[c]{@{}c@{}}  6.922 \\ 0.685 \end{tabular} & 
    \begin{tabular}[c]{@{}c@{}}  1.104 \\ 0.168 \end{tabular} \\
    \cline{2-7}
& INHA  & 
    \begin{tabular}[c]{@{}c@{}} -\\ - \end{tabular} & 
    \begin{tabular}[c]{@{}c@{}} -\\ - \end{tabular} & 
    \begin{tabular}[c]{@{}c@{}} -\\ - \end{tabular} &
    \begin{tabular}[c]{@{}c@{}} -\\ - \end{tabular} & 
    \begin{tabular}[c]{@{}c@{}} 5.716 \\ 0.484 \end{tabular} \\
    \hline
    \hline
\multirow{8}{*}{\begin{turn}{90}straight I (Lap 1)\end{turn}}
& KAIST & 
    \begin{tabular}[c]{@{}c@{}} - (KLD)\\ - (Mean) \end{tabular} & 
    \begin{tabular}[c]{@{}c@{}}  10.934 \\ 1.374 \end{tabular} & 
    \begin{tabular}[c]{@{}c@{}}  8.885 \\ 1.391 \end{tabular} & 
    \begin{tabular}[c]{@{}c@{}}  21.492 \\ 4.903 \end{tabular} & 
    \begin{tabular}[c]{@{}c@{}}  9.286 \\ 1.424 \end{tabular} \\
    \cline{2-7}
& CBNU  & 
    \begin{tabular}[c]{@{}c@{}} -\\ - \end{tabular} & 
    \begin{tabular}[c]{@{}c@{}} -\\ - \end{tabular} & 
    \begin{tabular}[c]{@{}c@{}}  1.813 \\ 0.338 \end{tabular} & 
    \begin{tabular}[c]{@{}c@{}}  11.000 \\ 3.706 \end{tabular} & 
    \begin{tabular}[c]{@{}c@{}}  1.501 \\ 0.176 \end{tabular} \\
    \cline{2-7}
& INU   & 
    \begin{tabular}[c]{@{}c@{}} -\\ - \end{tabular} & 
    \begin{tabular}[c]{@{}c@{}} -\\ - \end{tabular} & 
    \begin{tabular}[c]{@{}c@{}} -\\ - \end{tabular} &
    \begin{tabular}[c]{@{}c@{}}  12.540 \\ 3.643 \end{tabular} & 
    \begin{tabular}[c]{@{}c@{}}  0.302 \\ 0.163 \end{tabular} \\
    \cline{2-7}
& INHA  & 
    \begin{tabular}[c]{@{}c@{}} -\\ - \end{tabular} & 
    \begin{tabular}[c]{@{}c@{}} -\\ - \end{tabular} & 
    \begin{tabular}[c]{@{}c@{}} -\\ - \end{tabular} &
    \begin{tabular}[c]{@{}c@{}} -\\ - \end{tabular} & 
    \begin{tabular}[c]{@{}c@{}}  12.303 \\ 3.570 \end{tabular} \\
    \hline
    \hline
\multirow{8}{*}{\begin{turn}{90}straight I (Lap 2)\end{turn}}
& KAIST & 
    \begin{tabular}[c]{@{}c@{}} - (KLD)\\ - (Mean) \end{tabular} & 
    \begin{tabular}[c]{@{}c@{}}  9.045 \\ 1.354 \end{tabular} & 
    \begin{tabular}[c]{@{}c@{}}  21.421 \\ 5.142 \end{tabular} & 
    \begin{tabular}[c]{@{}c@{}}  13.283 \\ 2.089 \end{tabular} & 
    \begin{tabular}[c]{@{}c@{}}  16.984 \\ 3.881 \end{tabular} \\
    \cline{2-7}
& CBNU  & 
    \begin{tabular}[c]{@{}c@{}} -\\ - \end{tabular} & 
    \begin{tabular}[c]{@{}c@{}} -\\ - \end{tabular} & 
    \begin{tabular}[c]{@{}c@{}}  12.204 \\ 5.161 \end{tabular} & 
    \begin{tabular}[c]{@{}c@{}}  4.156 \\ 1.709 \end{tabular} & 
    \begin{tabular}[c]{@{}c@{}}  7.866 \\ 2.496 \end{tabular} \\
    \cline{2-7}
& INU   & 
    \begin{tabular}[c]{@{}c@{}} -\\ - \end{tabular} & 
    \begin{tabular}[c]{@{}c@{}} -\\ - \end{tabular} & 
    \begin{tabular}[c]{@{}c@{}} -\\ - \end{tabular} &
    \begin{tabular}[c]{@{}c@{}} 8.321 \\ 3.550 \end{tabular} & 
    \begin{tabular}[c]{@{}c@{}} 4.307 \\ 4.976 \end{tabular} \\
    \cline{2-7}
& INHA  & 
    \begin{tabular}[c]{@{}c@{}} -\\ - \end{tabular} & 
    \begin{tabular}[c]{@{}c@{}} -\\ - \end{tabular} & 
    \begin{tabular}[c]{@{}c@{}} -\\ - \end{tabular} &
    \begin{tabular}[c]{@{}c@{}} -\\ - \end{tabular} & 
    \begin{tabular}[c]{@{}c@{}}  3.773 \\ 1.700 \end{tabular} \\
    \hline
\end{tabular}
\end{table}

\subsection{Traffic analysis}
\begin{figure*}[ht]
    \centering
    \includegraphics[width=2.0\columnwidth]{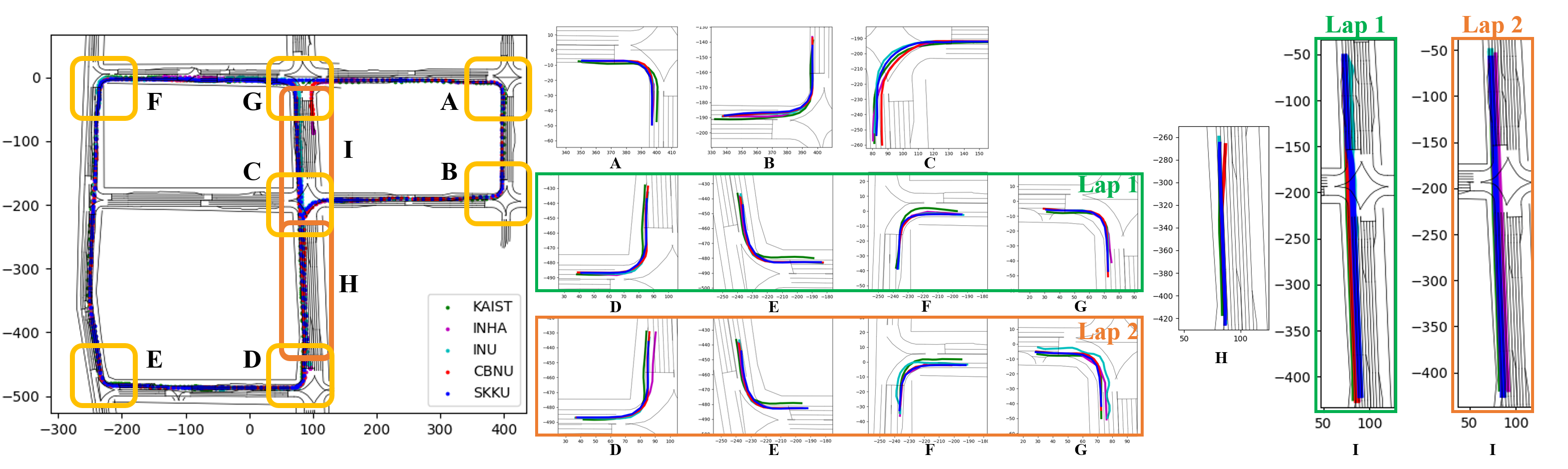}
    \caption{The result of the positioning log data are depicted. We could obtain real-world data from all the autonomously driven vehicles.
    Therefore, we could extract the similarities between the five teams in terms of spatial factors. 
    In addition, we separated the situations in terms of curve(Area : A,B, D-G), intersection(Area : C), and straight road(Area : H,I).}
    \label{fig:traffic_result_1}
\end{figure*}

Traffic analysis was conducted by comparing the trajectories of participating teams. 
Vehicle position data were collected from the location information every team sent to the infrastructure for traffic signal recognition. One of the six teams had technical issue in location feed, and is excluded in the analysis. 
The result of the positioning log data are shown in Fig. \ref{fig:traffic_result_1}.
In addition, we conducted section-wise analyses including intersection, curve, and straight section. 
We found that curve and intersection sections are the most critical component to win the competition, as the efficient maneuver in the congested in the corners and intersections greatly affect the overall performance.
\\
The results of the spatial similarity between the teams are presented in Table \ref{tab:full_race_traffic} using Eqs. \eqref{eq_spatial_kl_divergence_r_and_s} and \eqref{eq_mean_distance_error}. 
From the Table \ref{tab:full_race_traffic}, one can observe that traffic data from our team (KAIST) differed from other team data, owing to whether GPS was utilized in the localization method ---i.e., because the global position was transformed from our 3-D pointcloud map without using GPS, a continuous error appeared for the entire route.
However, when comparing teams other than ours with each other, both the KLD and RMSE values indicate high similarities in routes for the entire race.
This result implies that every team is likely to utilize a route related to obeying traffic laws, minimizing distance, and minimizing lane changes. 
The section-specific comparisons are presented in Tables \ref{tab:curve_1} - \ref{tab:straight}.
In addition, when the number of spatial data is insufficient to compare data distributions, the similarity between teams was quantified using RMSE rather than KLD.

\section{Conclusion}
\label{sec:conclusion}
In this paper, we proposed a full autonomous driving software stack to deploy a competitive driving model covering module-wise autonomous driving modules. 
In addition, autonomous modules consisting of navigation, perception, and planning systems were developed to enable our autonomous vehicle to be deployed in a complex urban area. 
We evaluated our system in the real world and won a competition for full-scale autonomous vehicles in South Korea. 
The traffic analyses provided additional insight on a multi-agent autonomous vehicle driving environment. 
The similarity in the projected optimal route of individual teams created conflicts among vehicles, which resulted in the performance loss. 
The strong performance of our architecture demonstrated the importance of adaptable approach in non-cooperative multi-agent autonomous driving. 

\bibliographystyle{unsrt}
\bibliography{citation}

\begin{thebibliography}{10}

\bibitem{morales2020automated}
Walter Morales-Alvarez, Oscar Sipele, R{\'e}gis L{\'e}beron, Hadj~Hamma
  Tadjine, and Cristina Olaverri-Monreal.
\newblock Automated driving: A literature review of the take over request in
  conditional automation.
\newblock {\em Electronics}, 9(12):2087, 2020.

\bibitem{fagnant2015preparing}
Daniel~J Fagnant and Kara Kockelman.
\newblock Preparing a nation for autonomous vehicles: opportunities, barriers
  and policy recommendations.
\newblock {\em Transportation Research Part A: Policy and Practice},
  77:167--181, 2015.

\bibitem{shim2015autonomous}
Inwook Shim, Jongwon Choi, Seunghak Shin, Tae-Hyun Oh, Unghui Lee, Byungtae
  Ahn, Dong-Geol Choi, David~Hyunchul Shim, and In-So Kweon.
\newblock An autonomous driving system for unknown environments using a unified
  map.
\newblock {\em IEEE transactions on intelligent transportation systems},
  16(4):1999--2013, 2015.

\bibitem{jo2013overall}
Kichun Jo, Minchae Lee, Dongchul Kim, Junsoo Kim, Chulhoon Jang, Euiyun Kim,
  Sangkwon Kim, Donghwi Lee, Changsup Kim, Seungki Kim, et~al.
\newblock Overall reviews of autonomous vehicle a1-system architecture and
  algorithms.
\newblock {\em IFAC Proceedings Volumes}, 46(10):114--119, 2013.

\bibitem{lee2016eurecar}
Unghui Lee, Jiwon Jung, Seunghak Shin, Yongseop Jeong, Kibaek Park,
  David~Hyunchul Shim, and In-so Kweon.
\newblock Eurecar turbo: A self-driving car that can handle adverse weather
  conditions.
\newblock In {\em 2016 IEEE/RSJ International Conference on Intelligent Robots
  and Systems (IROS)}, pages 2301--2306. IEEE, 2016.

\bibitem{jung2020v2x}
Chanyoung Jung, Daegyu Lee, Seungwook Lee, and David~Hyunchul Shim.
\newblock V2x-communication-aided autonomous driving: system design and
  experimental validation.
\newblock {\em Sensors}, 20(10):2903, 2020.

\bibitem{bojarski2016end}
Mariusz Bojarski, Davide Del~Testa, Daniel Dworakowski, Bernhard Firner, Beat
  Flepp, Prasoon Goyal, Lawrence~D Jackel, Mathew Monfort, Urs Muller, Jiakai
  Zhang, et~al.
\newblock End to end learning for self-driving cars.
\newblock {\em arXiv preprint arXiv:1604.07316}, 2016.

\bibitem{codevilla2018end}
Felipe Codevilla, Matthias M{\"u}ller, Antonio L{\'o}pez, Vladlen Koltun, and
  Alexey Dosovitskiy.
\newblock End-to-end driving via conditional imitation learning.
\newblock In {\em 2018 IEEE international conference on robotics and automation
  (ICRA)}, pages 4693--4700. IEEE, 2018.

\bibitem{xiao2020multimodal}
Yi~Xiao, Felipe Codevilla, Akhil Gurram, Onay Urfalioglu, and Antonio~M
  L{\'o}pez.
\newblock Multimodal end-to-end autonomous driving.
\newblock {\em IEEE Transactions on Intelligent Transportation Systems}, 2020.

\bibitem{kato2018autoware}
Shinpei Kato, Shota Tokunaga, Yuya Maruyama, Seiya Maeda, Manato Hirabayashi,
  Yuki Kitsukawa, Abraham Monrroy, Tomohito Ando, Yusuke Fujii, and Takuya
  Azumi.
\newblock Autoware on board: Enabling autonomous vehicles with embedded
  systems.
\newblock In {\em 2018 ACM/IEEE 9th International Conference on Cyber-Physical
  Systems (ICCPS)}, pages 287--296. IEEE, 2018.

\bibitem{fan2018baidu}
Haoyang Fan, Fan Zhu, Changchun Liu, Liangliang Zhang, Li~Zhuang, Dong Li,
  Weicheng Zhu, Jiangtao Hu, Hongye Li, and Qi~Kong.
\newblock Baidu apollo em motion planner.
\newblock {\em arXiv preprint arXiv:1807.08048}, 2018.

\bibitem{raju2019performance}
Vysyaraju~Manikanta Raju, Vrinda Gupta, and Shailesh Lomate.
\newblock Performance of open autonomous vehicle platforms: Autoware and
  apollo.
\newblock In {\em 2019 IEEE 5th International Conference for Convergence in
  Technology (I2CT)}, pages 1--5. IEEE, 2019.

\bibitem{pendleton2017perception}
Scott~Drew Pendleton, Hans Andersen, Xinxin Du, Xiaotong Shen, Malika Meghjani,
  You~Hong Eng, Daniela Rus, and Marcelo~H Ang~Jr.
\newblock Perception, planning, control, and coordination for autonomous
  vehicles.
\newblock {\em Machines}, 5(1):6, 2017.

\bibitem{dai2020perception}
Yanyan Dai and Suk-Gyu Lee.
\newblock Perception, planning and control for self-driving system based on
  on-board sensors.
\newblock {\em Advances in Mechanical Engineering}, 12(9):1687814020956494,
  2020.

\bibitem{tokunaga2019idf}
Shota Tokunaga, Yuki Horita, Yasuhiro Oda, and Takuya Azumi.
\newblock Idf-autoware: Integrated development framework for ros-based
  self-driving systems using matlab/simulink.
\newblock In {\em Workshop on Autonomous Systems Design (ASD 2019)}. Schloss
  Dagstuhl-Leibniz-Zentrum fuer Informatik, 2019.

\bibitem{zang2022winning}
Zirui Zang, Renukanandan Tumu, Johannes Betz, Hongrui Zheng, and Rahul
  Mangharam.
\newblock Winning the 3rd japan automotive ai challenge--autonomous racing with
  the autoware. auto open source software stack.
\newblock {\em arXiv preprint arXiv:2206.00770}, 2022.

\bibitem{thrun2006stanley}
Sebastian Thrun, Mike Montemerlo, Hendrik Dahlkamp, David Stavens, Andrei Aron,
  James Diebel, Philip Fong, John Gale, Morgan Halpenny, Gabriel Hoffmann,
  et~al.
\newblock Stanley: The robot that won the darpa grand challenge.
\newblock {\em Journal of field Robotics}, 23(9):661--692, 2006.

\bibitem{agha2021nebula}
Ali Agha, Kyohei Otsu, Benjamin Morrell, David~D Fan, Rohan Thakker, Angel
  Santamaria-Navarro, Sung-Kyun Kim, Amanda Bouman, Xianmei Lei, Jeffrey
  Edlund, et~al.
\newblock Nebula: Quest for robotic autonomy in challenging environments; team
  costar at the darpa subterranean challenge.
\newblock {\em arXiv preprint arXiv:2103.11470}, 2021.

\bibitem{song2015darpa}
Jia Song and Jim Alves-Foss.
\newblock The darpa cyber grand challenge: A competitor's perspective.
\newblock {\em IEEE Security \& Privacy}, 13(6):72--76, 2015.

\bibitem{lim2017robot}
Jeongsoo Lim, Inho Lee, Inwook Shim, Hyobin Jung, Hyun~Min Joe, Hyoin Bae,
  Okkee Sim, Jaesung Oh, Taejin Jung, Seunghak Shin, et~al.
\newblock Robot system of drc-hubo+ and control strategy of team kaist in darpa
  robotics challenge finals.
\newblock {\em Journal of Field Robotics}, 34(4):802--829, 2017.

\bibitem{o2020f1tenth}
Matthew O'Kelly, Hongrui Zheng, Dhruv Karthik, and Rahul Mangharam.
\newblock F1tenth: An open-source evaluation environment for continuous control
  and reinforcement learning.
\newblock {\em Proceedings of Machine Learning Research}, 123, 2020.

\bibitem{patton2021neuromorphic}
Robert Patton, Catherine Schuman, Shruti Kulkarni, Maryam Parsa, J~Parker
  Mitchell, N~Quentin Haas, Christopher Stahl, Spencer Paulissen, Prasanna
  Date, Thomas Potok, et~al.
\newblock Neuromorphic computing for autonomous racing.
\newblock In {\em International Conference on Neuromorphic Systems 2021}, pages
  1--5, 2021.

\bibitem{herrmann2020real}
Thomas Herrmann, Alexander Wischnewski, Leonhard Hermansdorfer, Johannes Betz,
  and Markus Lienkamp.
\newblock Real-time adaptive velocity optimization for autonomous electric cars
  at the limits of handling.
\newblock {\em IEEE Transactions on Intelligent Vehicles}, 2020.

\bibitem{jung2021game}
Chanyoung Jung, Seungwook Lee, Hyunki Seong, Andrea Finazzi, and David~Hyunchul
  Shim.
\newblock Game-theoretic model predictive control with data-driven
  identification of vehicle model for head-to-head autonomous racing.
\newblock {\em arXiv preprint arXiv:2106.04094}, 2021.

\bibitem{herrmann2020minimum}
Thomas Herrmann, Francesco Passigato, Johannes Betz, and Markus Lienkamp.
\newblock Minimum race-time planning-strategy for an autonomous electric
  racecar.
\newblock In {\em 2020 IEEE 23rd International Conference on Intelligent
  Transportation Systems (ITSC)}, pages 1--6. IEEE, 2020.

\bibitem{lee2022resilient}
Daegyu Lee, Chanyoung Jung, Andrea Finazzi, Hyunki Seong, and D~Hyunchul Shim.
\newblock Resilient navigation and path planning system for high-speed
  autonomous race car.
\newblock {\em arXiv preprint arXiv:2207.12232}, 2022.

\bibitem{quigley2009ros}
Morgan Quigley, Ken Conley, Brian Gerkey, Josh Faust, Tully Foote, Jeremy
  Leibs, Rob Wheeler, Andrew~Y Ng, et~al.
\newblock Ros: an open-source robot operating system.
\newblock In {\em ICRA workshop on open source software}, volume~3, page~5.
  Kobe, Japan, 2009.

\bibitem{Kos}
T.~{Kos}, I.~{Markezic}, and J.~{Pokrajcic}.
\newblock Effects of multipath reception on gps positioning performance.
\newblock In {\em Proceedings ELMAR-2010}, pages 399--402, 2010.

\bibitem{Welch}
Gary Bishop, Greg Welch, et~al.
\newblock An introduction to the kalman filter.
\newblock {\em Proc of SIGGRAPH, Course}, 8(27599-23175):41, 2001.

\bibitem{Honghui}
{Honghui Qi} and J.~B. {Moore}.
\newblock Direct kalman filtering approach for gps/ins integration.
\newblock {\em IEEE Transactions on Aerospace and Electronic Systems},
  38(2):687--693, 2002.

\bibitem{Reina}
G.~{Reina}, A.~{Vargas}, K.~{Nagatani}, and K.~{Yoshida}.
\newblock Adaptive kalman filtering for gps-based mobile robot localization.
\newblock In {\em 2007 IEEE International Workshop on Safety, Security and
  Rescue Robotics}, pages 1--6, 2007.

\bibitem{Hu}
Gaoge Hu, Bingbing Gao, Yongmin Zhong, and Chengfan Gu.
\newblock Unscented kalman filter with process noise covariance estimation for
  vehicular ins/gps integration system.
\newblock {\em Information Fusion}, 64:194--204, 2020.

\bibitem{larsen1998location}
Thomas~Dall Larsen, Martin Bak, Nils~A Andersen, and Ole Ravn.
\newblock Location estimation for an autonomously guided vehicle using an
  augmented kalman filter to autocalibrate the odometry.
\newblock In {\em FUSION98 Spie Conference}. Citeseer, 1998.

\bibitem{larsen1999design}
Thomas~Dall Larsen, Karsten~Lentfer Hansen, Nils~A Andersen, and Ole Ravn.
\newblock Design of kalman filters for mobile robots; evaluation of the
  kinematic and odometric approach.
\newblock In {\em Proceedings of the 1999 IEEE international conference on
  control applications (Cat. No. 99CH36328)}, volume~2, pages 1021--1026. IEEE,
  1999.

\bibitem{martinelli2003estimating}
Agostino Martinelli and Roland Siegwart.
\newblock Estimating the odometry error of a mobile robot during navigation.
\newblock In {\em 1st European Conference on Mobile Robots (ECMR 2003)}, number
  CONF, 2003.

\bibitem{ali2021real}
Rahmat Ali, Dongho Kang, Gahyun Suh, and Young-Jin Cha.
\newblock Real-time multiple damage mapping using autonomous uav and deep
  faster region-based neural networks for gps-denied structures.
\newblock {\em Automation in Construction}, 130:103831, 2021.

\bibitem{9328323}
Wanli Liu, Zhixiong Li, Shuaishuai Sun, Munish~Kumar Gupta, Haiping Du, Reza
  Malekian, Miguel~Angel Sotelo, and Weihua Li.
\newblock Design a novel target to improve positioning accuracy of autonomous
  vehicular navigation system in gps denied environments.
\newblock {\em IEEE Transactions on Industrial Informatics}, 17(11):7575--7588,
  2021.

\bibitem{lee2021assistive}
Daegyu Lee, Gyuree Kang, Boseong Kim, and D~Hyunchul Shim.
\newblock Assistive delivery robot application for real-world postal services.
\newblock {\em IEEE Access}, 9:141981--141998, 2021.

\bibitem{Caselitz}
T.~{Caselitz}, B.~{Steder}, M.~{Ruhnke}, and W.~{Burgard}.
\newblock Monocular camera localization in 3d lidar maps.
\newblock In {\em 2016 IEEE/RSJ International Conference on Intelligent Robots
  and Systems (IROS)}, pages 1926--1931, 2016.

\bibitem{Gu}
E.~{Javanmardi}, M.~{Javanmardi}, Y.~{Gu}, and S.~{Kamijo}.
\newblock Autonomous vehicle self-localization based on multilayer 2d vector
  map and multi-channel lidar.
\newblock In {\em 2017 IEEE Intelligent Vehicles Symposium (IV)}, pages
  437--442, 2017.

\bibitem{Javanmardi}
E.~{Javanmardi}, M.~{Javanmardi}, Y.~{Gu}, and S.~{Kamijo}.
\newblock Pre-estimating self-localization error of ndt-based map-matching from
  map only.
\newblock {\em IEEE Transactions on Intelligent Transportation Systems}, pages
  1--15, 2020.

\bibitem{biber2003normal}
Peter Biber and Wolfgang Stra{\ss}er.
\newblock The normal distributions transform: A new approach to laser scan
  matching.
\newblock In {\em Proceedings 2003 IEEE/RSJ International Conference on
  Intelligent Robots and Systems (IROS 2003)(Cat. No. 03CH37453)}, volume~3,
  pages 2743--2748. IEEE, 2003.

\bibitem{ulacs20133d}
Cihan Ula{\c{s}} and Hakan Temelta{\c{s}}.
\newblock 3d multi-layered normal distribution transform for fast and long
  range scan matching.
\newblock {\em Journal of Intelligent \& Robotic Systems}, 71(1):85--108, 2013.

\bibitem{LOAM}
Ji~Zhang and Sanjiv Singh.
\newblock Loam: Lidar odometry and mapping in real-time.
\newblock In {\em Robotics: Science and Systems}, volume~2, 2014.

\bibitem{lego}
Tixiao Shan and Brendan Englot.
\newblock Lego-loam: Lightweight and ground-optimized lidar odometry and
  mapping on variable terrain.
\newblock In {\em 2018 IEEE/RSJ International Conference on Intelligent Robots
  and Systems (IROS)}, pages 4758--4765. IEEE, 2018.

\bibitem{lio}
Tixiao Shan, Brendan Englot, Drew Meyers, Wei Wang, Carlo Ratti, and Daniela
  Rus.
\newblock Lio-sam: Tightly-coupled lidar inertial odometry via smoothing and
  mapping.
\newblock {\em arXiv preprint arXiv:2007.00258}, 2020.

\bibitem{xu2021fast}
Wei Xu and Fu~Zhang.
\newblock Fast-lio: A fast, robust lidar-inertial odometry package by
  tightly-coupled iterated kalman filter.
\newblock {\em IEEE Robotics and Automation Letters}, 6(2):3317--3324, 2021.

\bibitem{nguyen2021liro}
Thien-Minh Nguyen, Muqing Cao, Shenghai Yuan, Yang Lyu, Thien~Hoang Nguyen, and
  Lihua Xie.
\newblock Liro: Tightly coupled lidar-inertia-ranging odometry.
\newblock In {\em 2021 IEEE International Conference on Robotics and Automation
  (ICRA)}, pages 14484--14490. IEEE, 2021.

\bibitem{sommer2020openmp}
Lukas Sommer and Andreas Koch.
\newblock Openmp device offloading for embedded heterogeneous
  platforms-work-in-progress.
\newblock In {\em 2020 International Conference on Embedded Software (EMSOFT)},
  pages 4--6. IEEE, 2020.

\bibitem{dellenbach2022ct}
Pierre Dellenbach, Jean-Emmanuel Deschaud, Bastien Jacquet, and Fran{\c{c}}ois
  Goulette.
\newblock Ct-icp: Real-time elastic lidar odometry with loop closure.
\newblock In {\em 2022 International Conference on Robotics and Automation
  (ICRA)}, pages 5580--5586. IEEE, 2022.

\bibitem{chetverikov2002trimmed}
Dmitry Chetverikov, Dmitry Svirko, Dmitry Stepanov, and Pavel Krsek.
\newblock The trimmed iterative closest point algorithm.
\newblock In {\em Object recognition supported by user interaction for service
  robots}, volume~3, pages 545--548. IEEE, 2002.

\bibitem{rusu2009fast}
Radu~Bogdan Rusu, Nico Blodow, and Michael Beetz.
\newblock Fast point feature histograms (fpfh) for 3d registration.
\newblock In {\em 2009 IEEE international conference on robotics and
  automation}, pages 3212--3217. IEEE, 2009.

\bibitem{salti2014shot}
Samuele Salti, Federico Tombari, and Luigi Di~Stefano.
\newblock Shot: Unique signatures of histograms for surface and texture
  description.
\newblock {\em Computer Vision and Image Understanding}, 125:251--264, 2014.

\bibitem{koide2019portable}
Kenji Koide, Jun Miura, and Emanuele Menegatti.
\newblock A portable three-dimensional lidar-based system for long-term and
  wide-area people behavior measurement.
\newblock {\em International Journal of Advanced Robotic Systems},
  16(2):1729881419841532, 2019.

\bibitem{koide2021voxelized}
Kenji Koide, Masashi Yokozuka, Shuji Oishi, and Atsuhiko Banno.
\newblock Voxelized gicp for fast and accurate 3d point cloud registration.
\newblock In {\em 2021 IEEE International Conference on Robotics and Automation
  (ICRA)}, pages 11054--11059. IEEE, 2021.

\bibitem{gicp}
James Servos and Steven~L Waslander.
\newblock Multi-channel generalized-icp: A robust framework for multi-channel
  scan registration.
\newblock {\em Robotics and Autonomous systems}, 87:247--257, 2017.

\bibitem{kenney2011dedicated}
John~B Kenney.
\newblock Dedicated short-range communications (dsrc) standards in the united
  states.
\newblock {\em Proceedings of the IEEE}, 99(7):1162--1182, 2011.

\bibitem{park2011integrated}
Hyungjun Park, Adelin Miloslavov, Joyoung Lee, Malathi Veeraraghavan, Byungkyu
  Park, and Brian~Lee Smith.
\newblock Integrated traffic--communication simulation evaluation environment
  for intellidrive applications using sae j2735 message sets.
\newblock {\em Transportation research record}, 2243(1):117--126, 2011.

\bibitem{li2020rtm3d}
Peixuan Li, Huaici Zhao, Pengfei Liu, and Feidao Cao.
\newblock Rtm3d: Real-time monocular 3d detection from object keypoints for
  autonomous driving.
\newblock In {\em European Conference on Computer Vision}, pages 644--660.
  Springer, 2020.

\bibitem{yukai_yang_2020_4294717}
Yukai Yang.
\newblock {FastMOT: High-Performance Multiple Object Tracking Based on Deep
  SORT and KLT}, November 2020.

\bibitem{wang2018lanenet}
Ze~Wang, Weiqiang Ren, and Qiang Qiu.
\newblock Lanenet: Real-time lane detection networks for autonomous driving.
\newblock {\em arXiv preprint arXiv:1807.01726}, 2018.

\bibitem{pardo2006downscaling}
Eulogio Pardo-Ig{\'u}zquiza, Mario Chica-Olmo, and Peter~M Atkinson.
\newblock Downscaling cokriging for image sharpening.
\newblock {\em Remote Sensing of Environment}, 102(1-2):86--98, 2006.

\bibitem{duchovn2014path}
Franti{\v{s}}ek Ducho{\v{n}}, Andrej Babinec, Martin Kajan, Peter Be{\v{n}}o,
  Martin Florek, Tom{\'a}{\v{s}} Fico, and Ladislav Juri{\v{s}}ica.
\newblock Path planning with modified a star algorithm for a mobile robot.
\newblock {\em Procedia Engineering}, 96:59--69, 2014.

\bibitem{tarjan1972depth}
Robert Tarjan.
\newblock Depth-first search and linear graph algorithms.
\newblock {\em SIAM journal on computing}, 1(2):146--160, 1972.

\bibitem{stahl2019multilayer}
Tim Stahl, Alexander Wischnewski, Johannes Betz, and Markus Lienkamp.
\newblock Multilayer graph-based trajectory planning for race vehicles in
  dynamic scenarios.
\newblock In {\em 2019 IEEE Intelligent Transportation Systems Conference
  (ITSC)}, pages 3149--3154. IEEE, 2019.

\bibitem{a_star}
Peter~E Hart, Nils~J Nilsson, and Bertram Raphael.
\newblock A formal basis for the heuristic determination of minimum cost paths.
\newblock {\em IEEE transactions on Systems Science and Cybernetics},
  4(2):100--107, 1968.

\bibitem{a_star_variation}
Zahra Boroujeni, Daniel Goehring, Fritz Ulbrich, Daniel Neumann, and Raul
  Rojas.
\newblock Flexible unit a-star trajectory planning for autonomous vehicles on
  structured road maps.
\newblock In {\em 2017 IEEE international conference on vehicular electronics
  and safety (ICVES)}, pages 7--12. IEEE, 2017.

\bibitem{hybrid_a_star}
Dmitri Dolgov, Sebastian Thrun, Michael Montemerlo, and James Diebel.
\newblock Practical search techniques in path planning for autonomous driving.
\newblock {\em Ann Arbor}, 1001(48105):18--80, 2008.

\bibitem{hybrid_a_star_parking}
Saeid Sedighi, Duong-Van Nguyen, and Klaus-Dieter Kuhnert.
\newblock Guided hybrid a-star path planning algorithm for valet parking
  applications.
\newblock In {\em 2019 5th international conference on control, automation and
  robotics (ICCAR)}, pages 570--575. IEEE, 2019.

\bibitem{paden2016survey}
Brian Paden, Michal {\v{C}}{\'a}p, Sze~Zheng Yong, Dmitry Yershov, and Emilio
  Frazzoli.
\newblock A survey of motion planning and control techniques for self-driving
  urban vehicles.
\newblock {\em IEEE Transactions on intelligent vehicles}, 1(1):33--55, 2016.

\bibitem{rrt_original}
Steven~M LaValle et~al.
\newblock Rapidly-exploring random trees: A new tool for path planning.
\newblock 1998.

\bibitem{rrt_darpa}
Yoshiaki Kuwata, Gaston~A Fiore, Justin Teo, Emilio Frazzoli, and Jonathan~P
  How.
\newblock Motion planning for urban driving using rrt.
\newblock In {\em 2008 IEEE/RSJ International Conference on Intelligent Robots
  and Systems}, pages 1681--1686. IEEE, 2008.

\bibitem{rrt_star_original}
Sertac Karaman and Emilio Frazzoli.
\newblock Sampling-based algorithms for optimal motion planning.
\newblock {\em The international journal of robotics research}, 30(7):846--894,
  2011.

\bibitem{informed_rrt_star}
Jonathan~D Gammell, Siddhartha~S Srinivasa, and Timothy~D Barfoot.
\newblock Informed rrt*: Optimal sampling-based path planning focused via
  direct sampling of an admissible ellipsoidal heuristic.
\newblock In {\em 2014 IEEE/RSJ International Conference on Intelligent Robots
  and Systems}, pages 2997--3004. IEEE, 2014.

\bibitem{rrt_exploration_mobile}
Hassan Umari and Shayok Mukhopadhyay.
\newblock Autonomous robotic exploration based on multiple rapidly-exploring
  randomized trees.
\newblock In {\em 2017 IEEE/RSJ International Conference on Intelligent Robots
  and Systems (IROS)}, pages 1396--1402. IEEE, 2017.

\bibitem{rrt_exploration_drone}
Andreas Bircher, Mina Kamel, Kostas Alexis, Helen Oleynikova, and Roland
  Siegwart.
\newblock Receding horizon path planning for 3d exploration and surface
  inspection.
\newblock {\em Autonomous Robots}, 42(2):291--306, 2018.

\bibitem{kinodynamic_rrt_star}
Dustin~J Webb and Jur Van Den~Berg.
\newblock Kinodynamic rrt*: Asymptotically optimal motion planning for robots
  with linear dynamics.
\newblock In {\em 2013 IEEE international conference on robotics and
  automation}, pages 5054--5061. IEEE, 2013.

\bibitem{sampling_with_constraints}
Zachary Kingston, Mark Moll, and Lydia~E Kavraki.
\newblock Sampling-based methods for motion planning with constraints.
\newblock {\em Annual review of control, robotics, and autonomous systems},
  1:159--185, 2018.

\bibitem{werling2010optimal}
Moritz Werling, Julius Ziegler, S{\"o}ren Kammel, and Sebastian Thrun.
\newblock Optimal trajectory generation for dynamic street scenarios in a
  frenet frame.
\newblock In {\em 2010 IEEE International Conference on Robotics and
  Automation}, pages 987--993. IEEE, 2010.

\bibitem{zhu2020trajectory}
Sheng Zhu and Bilin Aksun-Guvenc.
\newblock Trajectory planning of autonomous vehicles based on parameterized
  control optimization in dynamic on-road environments.
\newblock {\em Journal of Intelligent \& Robotic Systems}, 100(3):1055--1067,
  2020.

\bibitem{zheng2020bezier}
Ling Zheng, Pengyun Zeng, Wei Yang, Yinong Li, and Zhenfei Zhan.
\newblock B{\'e}zier curve-based trajectory planning for autonomous vehicles
  with collision avoidance.
\newblock {\em IET Intelligent Transport Systems}, 14(13):1882--1891, 2020.

\bibitem{conte2017elementary}
Samuel~Daniel Conte and Carl De~Boor.
\newblock {\em Elementary numerical analysis: an algorithmic approach}.
\newblock SIAM, 2017.

\bibitem{perez2008kullback}
Fernando P{\'e}rez-Cruz.
\newblock Kullback-leibler divergence estimation of continuous distributions.
\newblock In {\em 2008 IEEE international symposium on information theory},
  pages 1666--1670. IEEE, 2008.

\end{thebibliography}
\section{Appendix}
\begin{figure}[ht]
    \centering
    \includegraphics[width=1\columnwidth]{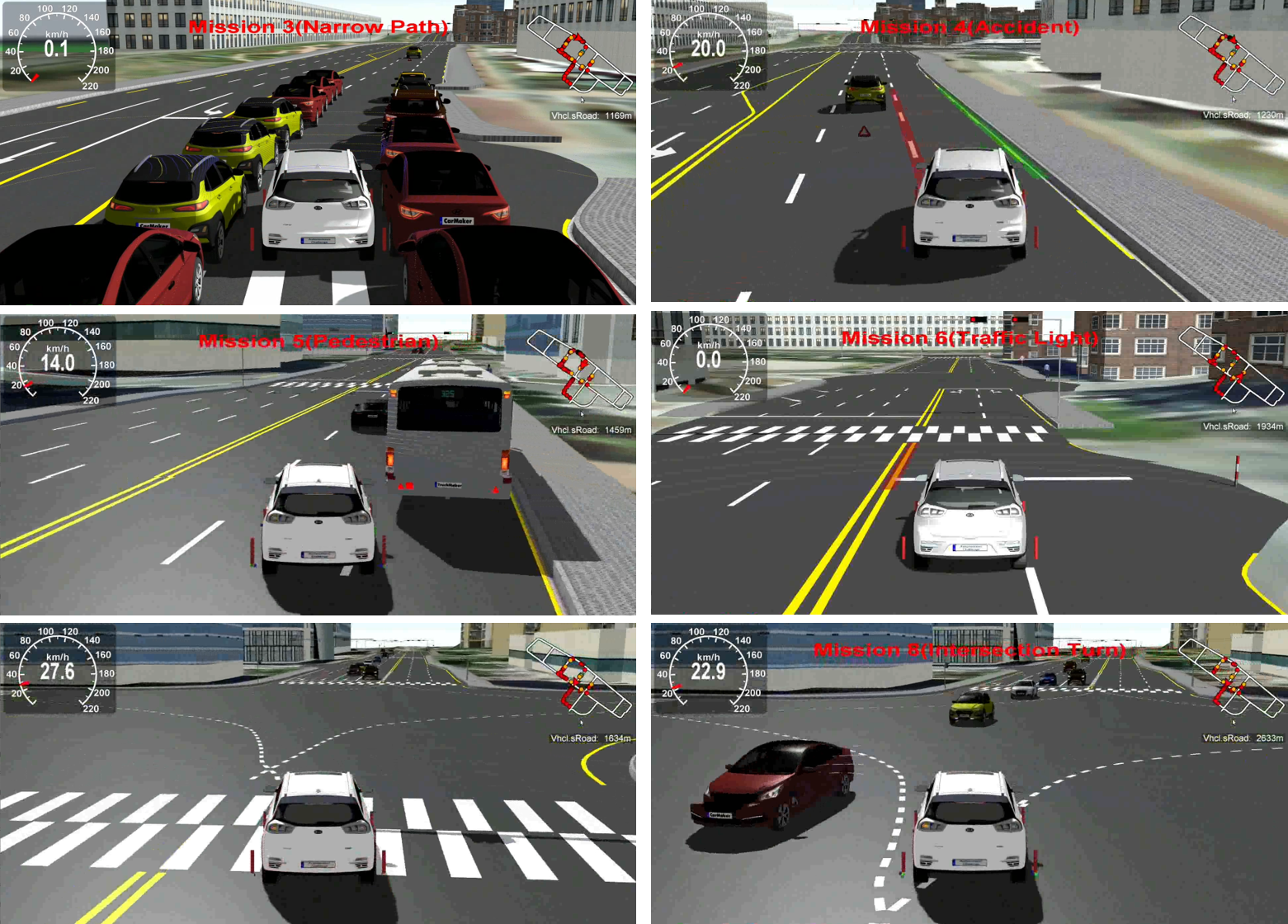}
    \caption{Simulation screen is depicted. The simulation qualification was a timed trial composed of eight missions: overtaking a low-speed vehicle, handling a cut-in vehicle, passing a narrow environment, accident vehicle avoidance, passing un-signalized intersection, and following the traffic signals.}
    \label{fig:appendix_1}
\end{figure}

This appendix provides additional results of the competition for simulation qualification and real-world competition. 
23 teams competed for simulation qualification, and only six teams qualified for final.
The simulation qualification was a timed trial composed of eight missions: overtaking a low-speed vehicle, handling a cut-in vehicle, passing a narrow passage, accident vehicle avoidance, passing un-signalized intersection, and following the traffic signals, as illustrated in Fig. \ref{fig:appendix_1}. \\
After qualification, the six qualifying teams were supported by Hyundai Motors providing an electronic vehicle, the KIA Niro. 
On November 29\textsuperscript{th}, 2021, the main event was held in Sangam, Seoul, which is a complex area where media and broadcasting companies are located in Seoul, and there are many steel structures and high-rise buildings, causing a GPS-degraded environment also known as urban canyons. 
In the final competition, our team won with the minimum transverse time. 
While driving autonomously, our vehicle was not penalized due to traffic laws or threatening driving. 
In addition, our vehicle overtook other teams four times, while detecting a low-speed vehicle and congested traffic scenarios.
\begin{figure}[ht]
    \centering
    \includegraphics[width=1\columnwidth]{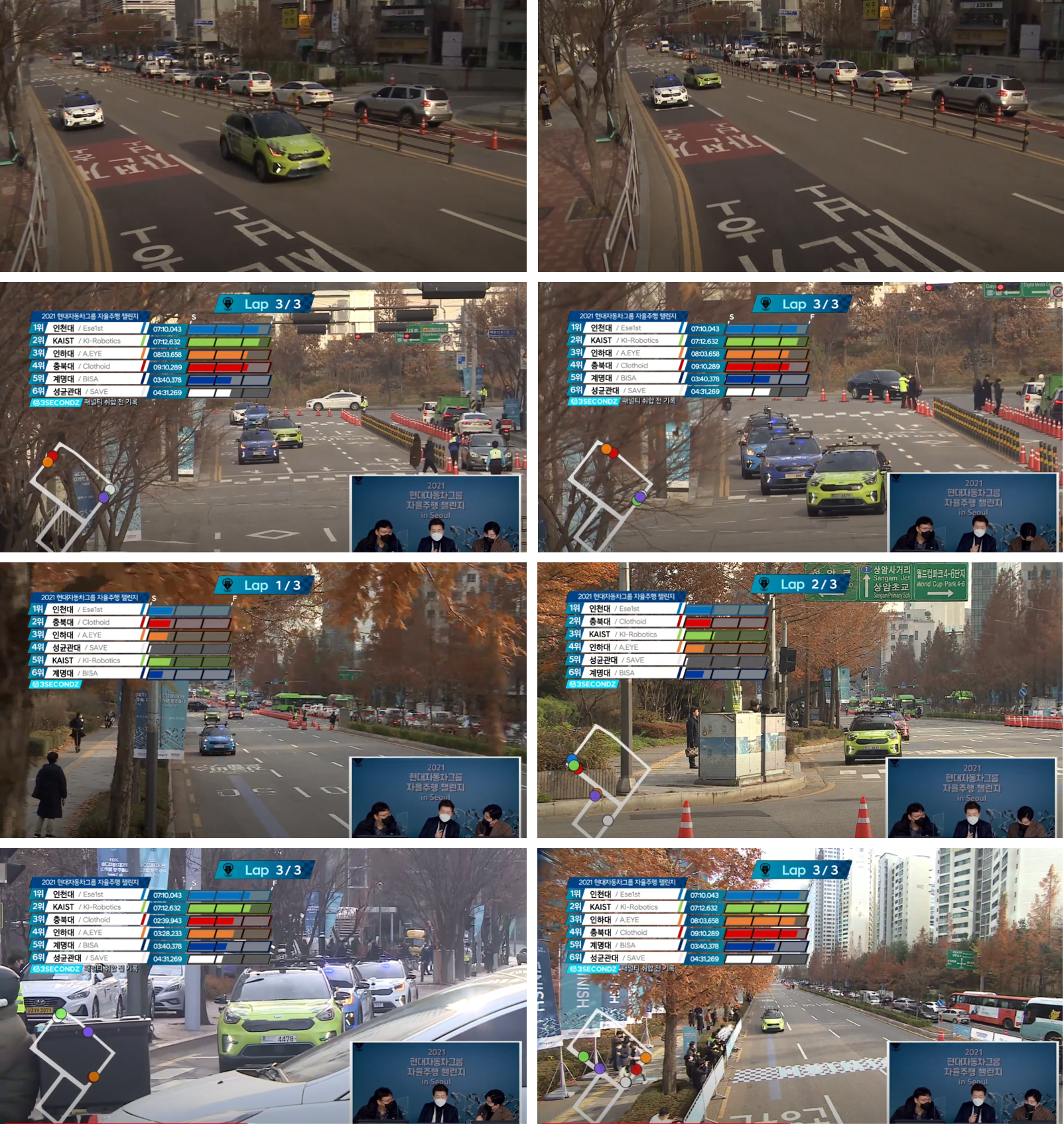}
    \caption{During the real-world final competition, multiple autonomous vehicles raced against each other. To win this competition, overtaking a low-speed vehicle, handling a cut-in vehicle, and following traffic signals should be developed by each team.}
    \label{fig:appendix_2}
\end{figure}
\begin{table}[hbt!]
\caption[final race result]{final race result}
\label{tab:final_race_result}
\centering
\begin{tabular}{c|cc|c}
\hline
Team & Lap time(min.'sec.") & Traffic violation & rank\\ \hline
\textbf{KAIST} & 
    \textbf{11' 27"} & 
    \textbf{0} &
    \textbf{1} \\
    \hline
CBNU  & 
    13' 31" & 
    0 &
    2\\
    \hline
INU   & 
    14' 19" & 
    0 &
    3 \\
    \hline
INHA  & 
    12' 31" & 
    1 &
    4 \\
    \hline
SKKU  & 
    15' 53" & 
    1 &
    5 \\
    \hline
KMU  & 
    19' 02" & 
    7 &
    6 \\
\hline
\end{tabular}
\end{table}

The results of the final event are depicted in Fig. \ref{fig:appendix_2} and listed in Table \ref{tab:final_race_result}. 
To enhance the safety of the autonomous vehicle, there were critical penalties for traffic violations in the final event. Moreover, the full competition video is available at \url{https://youtu.be/EJD34qMe768}.

\begin{IEEEbiography}[{\includegraphics[width=1in,height=1.25in,clip,keepaspectratio]{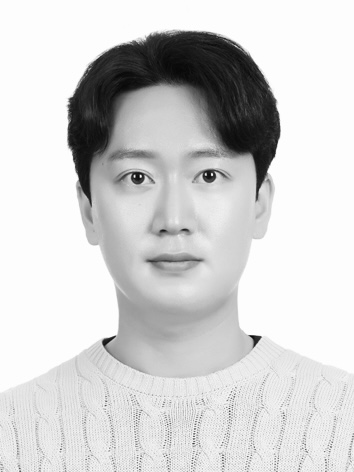}}]{Daegyu Lee} received his B.S. degree in Automotive 
Engineering from Kookmin University, Seoul, South Korea, in 2018, 
and his M.S. degree in Division of Future Vehicle from the Korea Advanced Institute of 
Science and Technology (KAIST), Daejeon, South Korea, in 2020. He is currently pursuing a Ph.D. degree in Electrical Engineering at KAIST.

His research interests include autonomous systems, robotics, motion planning, and localization based on unmanned ground vehicles.
\end{IEEEbiography}

\begin{IEEEbiography}[{\includegraphics[width=1in,height=1.25in,clip,keepaspectratio]{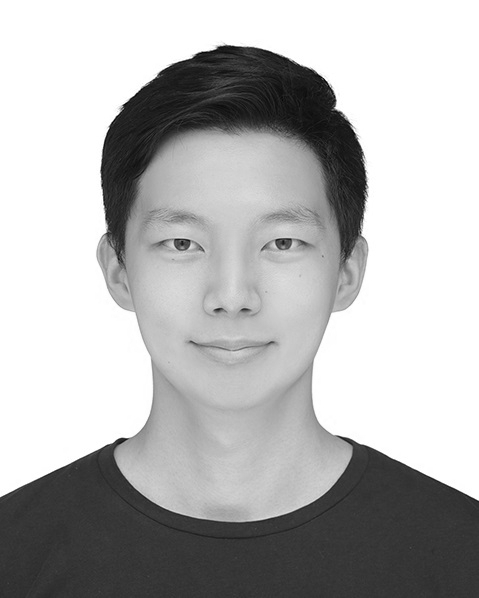}}]{Hyunki Seong} received his B.S. degree in Mechanical Engineering from Inha University, Incheon, South Korea, in 2019, and his M.S. degree in Division of Robotics from the Korea Advanced Institute of Science and Technology (KAIST), Daejeon, South Korea, in 2021. He is currently pursuing a Ph.D. degree in Electrical Engineering at KAIST.
His research interests include autonomous systems, robotics, and motion planning based on unmanned ground vehicles.
\end{IEEEbiography}

\begin{IEEEbiography}[{\includegraphics[width=1in,height=1.25in,clip,keepaspectratio]{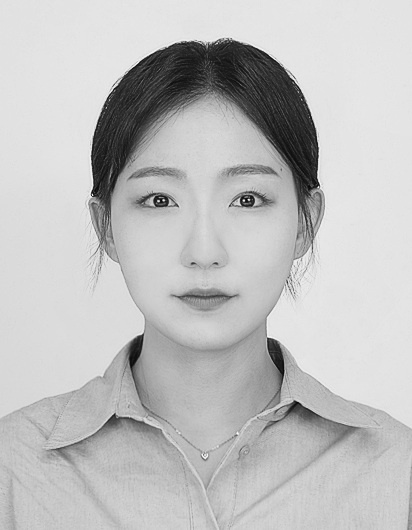}}]{Gyuree Kang} received her B.S. degree in Mechanical Engineering from Sungkyunkwan University, Suwon, Gyeonggi, South Korea, in 2020, and her M.S. degree in the Robotics Program from the Korea Advanced Institute of Science and Technology (KAIST), Daejeon, South Korea, in 2022. She is currently pursuing a Ph.D. degree in Electrical Engineering at KAIST.

Her research interests include autonomous systems, robotics, and task and motion planning.
\end{IEEEbiography}

\begin{IEEEbiography}[{\includegraphics[width=1in,height=1.25in,clip,keepaspectratio]{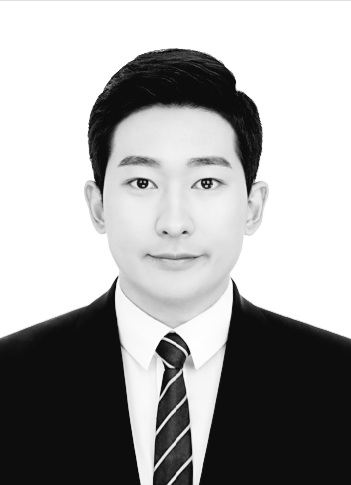}}]{Seungil Han} received his B.S. degree in Division of Robotics from Kwangwoon University, Seoul, South Korea, in 2020, and his M.S. degree in the Robotics Program from the Korea Advanced Institute of Science and Technology (KAIST), Daejeon, South Korea, in 2022. His research interests include autonomous systems, robotics, and motion planning based on unmanned ground vehicles.
\end{IEEEbiography}

\begin{IEEEbiography}[{\includegraphics[width=1in,height=1.25in,clip,keepaspectratio]{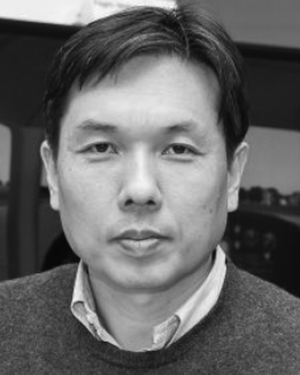}}]{D.Hyunchul Shim} received his B.S. and M.S. degrees in Mechanical Design and Production Engineering from Seoul National University, Seoul, South Korea, in 1991 and 1993, respectively, and his Ph.D. degree in Mechanical Engineering from the University of California at Berkeley, Berkeley, CA, USA, in 2000. He worked with the Hyundai Motor Company and Maxtor Corporation from 1993 to 1994 and from 2001 to 2005, respectively.
In 2007, he joined the Department of Aerospace Engineering, KAIST, Daejeon, South Korea, and is currently a tenured Professor with the Department of Electrical Engineering, and Adjunct Professor, Graduate School of AI, KAIST.
His research interests include control systems, autonomous vehicles, and robotics. He is also the Director of the Korea Civil RPAS Research Center.
\end{IEEEbiography}

\begin{IEEEbiography}[{\includegraphics[width=1in,height=1.25in,clip,keepaspectratio]{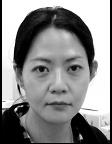}}]{Yoonjin Yoon} is an associate professor of Civil and Environmental Engineering at Korea Advanced Institute of Science and Technology (KAIST) with joint appointment in Graduate School Artificial Intelligence and Graduate School of Data Science. 
Her main research focus is computational transportation science. 
Some of her earlier works has dealt with stochastic geometric air traffic flow optimization, topological urban airspace analysis, urban energy optimization models. Her most recent efforts involve traffic forecast in urban mobility network using graph neural networks, and urban predictions using graph representation learning. 
She received B.S. in Mathematics from Seoul National University, dual M.S. degrees in Computer Science, and Management Science and Engineering from Stanford University. She received her Ph.D. in Civil and Environmental Engineering from University of California, Berkeley
\end{IEEEbiography}

\end{document}